%% file: main.tex
\documentclass{article}
\usepackage{iclr2027_conference,times}

\usepackage[utf8]{inputenc}
\usepackage[T1]{fontenc}
\usepackage{hyperref}
\usepackage{url}
\usepackage{booktabs}
\usepackage{amsfonts}
\usepackage{nicefrac}
\usepackage{microtype}
\usepackage{xcolor}
\usepackage{subcaption}
\usepackage{graphicx}
\usepackage{siunitx}
\usepackage{makecell}
\usepackage{amsmath}
\usepackage{multirow}
\usepackage{xcolor}
\usepackage{algorithm}
\usepackage{algpseudocode}
\usepackage{amssymb}

\algblockdefx[Except]{Except}{EndExcept}
  [1]{\textbf{except} #1\textbf{:}}
  {}
\algnotext{EndExcept}

\algrenewcommand\algorithmicdo{}
\algrenewcommand\algorithmicthen{}

\usepackage{threeparttable}
\usepackage[table]{xcolor}
\definecolor{cliffteal}{HTML}{00D2BE}
\definecolor{kimiblue}{HTML}{3B82F6}
\definecolor{glmred}{HTML}{FF2800}
\definecolor{citered}{HTML}{FF2800}
\definecolor{citepurple}{HTML}{AE8AFF}
\definecolor{citeteal}{HTML}{00D2BE}
\definecolor{yellow}{HTML}{FFDA33}

\hypersetup{
    colorlinks=true,
    citecolor=citeteal,
    linkcolor=citeteal,
    urlcolor=citeteal
}

\usepackage{tikz}
\definecolor{cliffink}{HTML}{1A1A1A}
\newcommand{\hb}[2][cliffink]{%
  \tikz[baseline=-0.55ex]{%
    \ifcase#2
      \draw[#1,line width=0.5pt] (0,0) circle (0.30em);
    \or
      \draw[#1,line width=0.5pt] (0,0) circle (0.30em);
      \fill[#1] (0,0) -- (90:0.30em) arc (90:270:0.30em) -- cycle;
    \or
      \fill[#1] (0,0) circle (0.31em);
    \fi}}
\newcommand{\yes}{\hb{2}}
\newcommand{\pa}{\hb{1}}
\newcommand{\no}{\hb{0}}

\title{\textsc{CliffCompaction}: Cost-Efficient Compaction \\ for Long-Horizon Coding Agents}

\author{
Trang Nguyen$^{1}$, Eulrang Cho$^{1}$, Bingqing Chen$^{2}$ \& Tim Dettmers$^{1}$ \\
$^{1}$Carnegie Mellon University,
\href{mailto:thientrangngv@cmu.edu,eulrang@cmu.edu,dettmers@cmu.edu}
{\texttt{\{thientrangngv,eulrang,dettmers\}@cmu.edu}} \\
$^{2}$Bosch Center for AI,
\href{mailto:bingqing.chen@us.bosch.com}
{\texttt{bingqing.chen@us.bosch.com}}
}

\iclrfinalcopy 

\begin{document}

\maketitle

\begin{abstract}

Agents often work on complex problems that require millions of tokens of context, which necessitates compacting across sessions due to limited context windows. We develop \textsc{CliffCompaction}, an autocompaction technique that reduces cost by up to 50\% under a bounded context while maintaining or improving performance on Terminal-Bench and achieving new levels of efficiency for test-time scaling and state-of-the-art results on KernelBench. The per-rollout savings of \textsc{CliffCompaction} make the performance--cost trade-off of test-time scaling more efficient, adding over 10 percentage points on Terminal-Bench for less than the cost of two full-context runs. Under parallel test-time scaling, \textsc{CliffCompaction} lets Kimi K2.6 match Opus 4.7, and exceed Opus 4.6 and GPT-5.3 Codex at lower cost. The key to ~\textsc{CliffCompaction}'s effectiveness is that it keeps compacted information faithful by only truncating or dropping content, never rephrasing or rewriting it. We never compact a compaction---each pass operates only on original content, and prior compacted output is discarded, preventing context drift from accumulating. These properties sustain continual learning over sessions exceeding a million tokens: on KernelBench, \textsc{CliffCompaction} reaches CUDA kernel speedups of $2.23\times$ after 200 steps and $3.58\times$ after 400 steps, surpassing specialized search algorithms and trained agents despite being a general-purpose compaction technique. We open-source a scaffold-agnostic API-proxy implementation of \textsc{CliffCompaction} usable with Claude Code, Codex and other harnesses.\footnote{\url{https://github.com/nguyenvuthientrang/cliffcompaction}}

\end{abstract}

 \begin{figure}[ht]
  \centering
  \includegraphics[width=\linewidth]{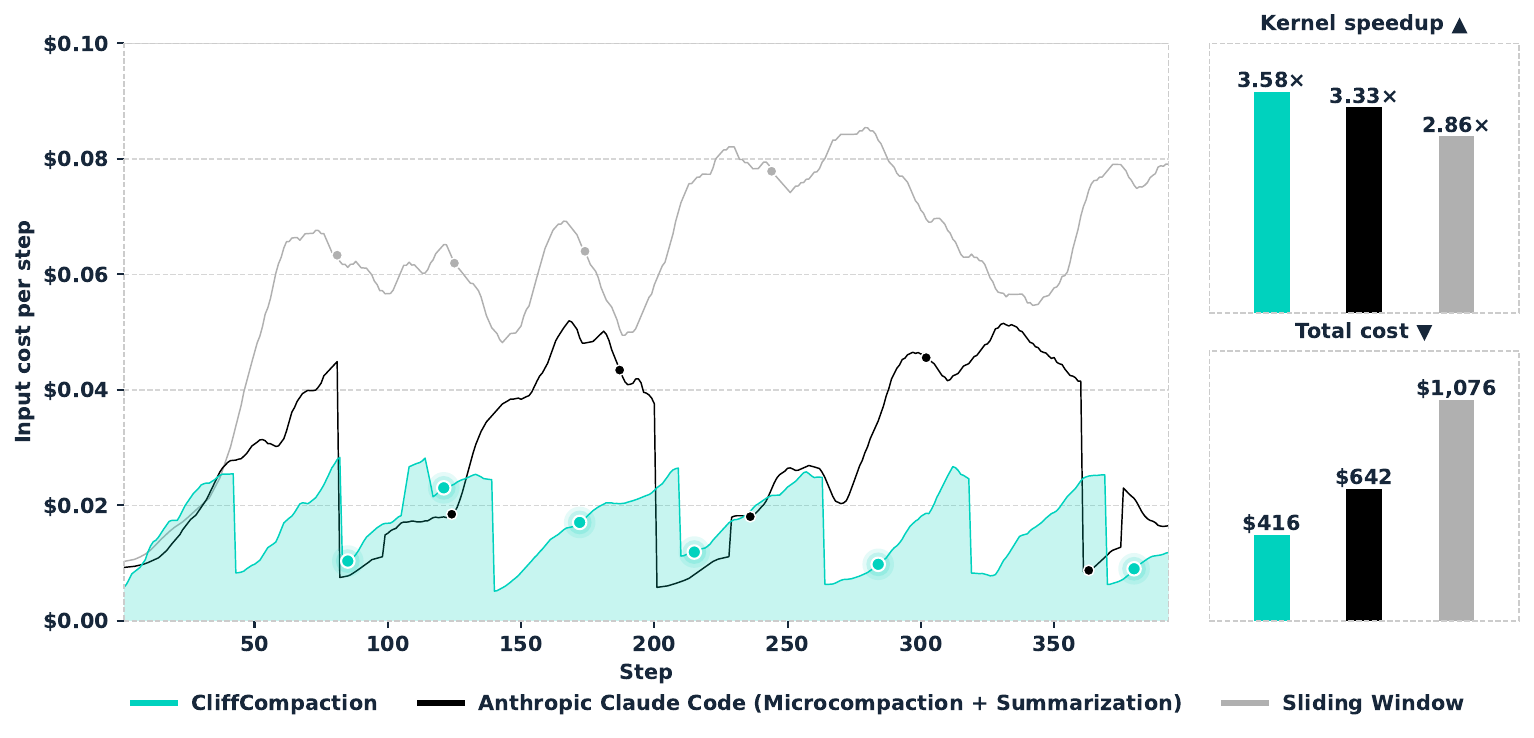}
  \caption{Continual learning of CUDA kernel development on KernelBench using \texttt{OpenHands}: \textbf{\textsc{CliffCompaction} is both the cheapest and the highest performing.}. \emph{Left:} per-step input cost on one Level~3 problem; dots mark when each method's overall speedup across 50 problems first reaches $1.0\times, 1.5\times, \ldots, 3.5\times$. \emph{Right:} final speedup ($\blacktriangle$) and total cost ($\blacktriangledown$) across all 50 problems.}
  \label{fig:compaction-input}
\end{figure}


\section{Introduction}
\input{contents/intro}

\section{\textsc{CliffCompaction}}

\input{contents/method}


\section{CliffCompaction Maintains Performance at Lower Cost}
\label{sec:main_experiment}
\input{contents/mainresults}


\section{Making Test-Time Scaling Affordable with \textsc{CliffCompaction}}
\vspace{-2pt}
\label{sec:tts}
\input{contents/tts}

\section{Continual Learning with \textsc{CliffCompaction}}
\input{contents/continuallearning}

\section{\textsc{CliffCompaction} vs. Existing Compaction Strategies}
\input{contents/compactioncomparison}

\section{Related Work}
\input{contents/relatedwork}

\section{Limitations}
\input{contents/conclusion}


\bibliography{references}
\bibliographystyle{iclr2027_conference}

\appendix

\section{Extended Efficiency Analysis}
\input{contents/efficiencyanalysis}
\input{contents/extended_efficiency}

\section{Extended \textsc{CliffCompaction} Design and Implementation}
\label{appendix:cliff}
\input{contents/extended_cliff}

\section{Extended Evaluation and Analysis}
\input{contents/extended_analysis}

\section{Experimental Details}
\input{contents/extended_setup}




\end{document}

%% file: contents/intro.tex
Coding agents solve complex tasks by interacting with tools and environments over contexts spanning millions of tokens. However, processing such long contexts is computationally expensive, and increasingly bloated contexts can impair agent effectiveness as trajectories grow longer. We show that bounded context with autocompaction can match or exceed full-context performance at lower cost.

We develop \textsc{CliffCompaction}, an autocompaction technique for coding agents. On SWE-bench Verified, \textsc{CliffCompaction} preserves full-context success rates for GLM-5.1 and Kimi K2.6 with context thresholds of only 32K and 16K tokens. On Terminal-Bench 2.0, it achieves higher success rates while reducing cost by 50\%.

For long context tasks, two main problems arise: (a) how to manage context information once an agent session runs out of context length; (b) how to maximize agent performance while reducing cost from long context.

For context management (a), the main approaches are either to {\it compact}---to truncate or summarize---the context or to store memories and start a new session with those memories. Sliding windows, which keep only the most recent turns or observations~\citep{yang2024sweagent}, are effective drop-in solutions, but invalidate the prefix cache whenever the window advances, increasing inference cost.
LLM-based summarization, though widely used, compounds loss across summary-of-summary chains.


A more demanding challenge in this area is continual learning, where an agent needs to continually improve previous solutions to a problem where the total context can reach millions of tokens. While solutions for continual learning exist, they often rely on complex, task-specific methods~\citep{du2026adaexplore, dai2026cuda} with limited generalizability.

On the efficiency side (b), the long context is represented by the KV-cache. While the KV-cache leads to efficient inference, it is the main memory and computational bottleneck for long-context sessions and increases the latency and inference cost significantly. While system-level {\it backend} solutions such as FlashAttention~\citep{dao2022flashattention} or DeepSeek Sparse Attention~\citep{deepseekai2025deepseekv32pushingfrontieropen} reduce the KV-cache memory requirements directly, token-based {\it frontend} algorithms seek to reduce KV-cache costs indirectly and improve model performance by manipulating the tokens in the context.

One particular problem for frontend efficiency is test-time scaling~\citep{kwok2026llmasaverifiergeneralpurposeverificationframework, kim2026scaling}, where one seeks to use a model repeatedly and thus use additional tokens to achieve favorable cost-performance trade-offs compared to another model, for example, a frontier model. However, test-time scaling often requires tens of rollouts and incurs exorbitant inference costs. Moreover, as scaling gains saturate and parallel scaling introduces a selection bottleneck, it remains unclear whether the marginal improvements justify the expense.

\textsc{CliffCompaction} addresses these challenges: it maintains performance on coding and terminal tasks at lower cost, makes test-time scaling economical, and sets a new best result for continual learning on KernelBench under a matched evaluation setup.

\input{tables/facet}

\textsc{CliffCompaction} is a rule-based context-management method for long-running agents. \textsc{CliffCompaction} lets the context grow naturally until it reaches a predefined threshold, at which point it performs compaction and reduces the accumulated context. 
During compaction, \textsc{CliffCompaction} truncates components that dominate context length, primarily tool calls and tool outputs. Across successive compaction events, it does not stack or recursively compress previously compacted history. Instead, each compaction discards the previous compacted history and constructs a new compacted block from the current active context. Discarding past sessions prevents the agent to act on partial information and avoids context drift. Table~\ref{tab:comparison} summarizes the design space and highlights our deliberate sacrifice of full-history recall in exchange for a method that is training-free, drop-in, model-agnostic, cache-friendly, and free of auxiliary LLM calls.


With this design, \textsc{CliffCompaction} jointly addresses the context-capacity and efficiency challenges of long-horizon agents (Figure~\ref{fig:compaction-input}). Our experiments demonstrate its effectiveness across multiple benchmarks, scaffolds, and models. \textsc{CliffCompaction} preserves agent performance while cutting token usage, reducing cost from cache reads by up to 90\% and curtailing total cost accordingly. These savings make test-time scaling economically viable: on Terminal-Bench 2.0, multiple compacted Kimi K2.6 rollouts match Opus 4.7 and exceed Opus 4.6 and GPT-5.3 Codex at lower cost. Scaling to three rollouts gains $10.5$ points at only $1.9\times$ the cost of a single uncompacted Kimi run. For continual learning, \textsc{CliffCompaction} reaches the best results while requiring no external memory or task-specific design. Under matched settings, it outperforms methods built specifically for kernel optimization by $25\%$, despite being a general compaction strategy. With our best configuration, it reaches a $3.58\times$ speedup on KernelBench Level 3 across sessions exceeding a million tokens of context. At equal spend, \textsc{CliffCompaction} buys more performance than full-context settings, making it an effective and practical compaction technique for long-horizon coding agents.

%% file: tables/facet.tex
\vspace{-5pt}
\begin{table}[!htbp]
\centering
\caption{Context-management approaches.
\yes: supported; \no: not supported; \pa: partial or conditional.
\textsc{CliffCompaction} sacrifices only full-history recall, which we show is not needed (\S\ref{sec:main_experiment}).}
\label{tab:comparison}
\setlength{\tabcolsep}{6pt}
\renewcommand{\arraystretch}{1.2}
\resizebox{\textwidth}{!}{%
\begin{tabular}{l ccccccc}
\toprule
\textbf{Approach}
 & \makecell{Training-\\free}
 & \makecell{Drop-\\in}
 & \makecell{Model-\\agnostic}
 & \makecell{No aux.\\LLM call}
 & \makecell{Cache-\\friendly}
 & \makecell{High\\precision}
 & \makecell{Full-history\\recall} \\
\midrule
Sliding window          & \yes & \yes & \yes & \yes & \no  & \yes & \no  \\
LLM summarization       & \yes & \yes & \yes & \no  & \yes & \no  & \pa  \\
Learned compression     & \no  & \no  & \no  & \no  & \pa  & \no  & \pa  \\
Sub-agent decomposition & \pa  & \no  & \pa  & \no  & \pa  & \pa  & \pa  \\
External memory / RAG   & \yes & \no  & \yes & \yes & \pa  & \yes & \yes \\
Structure-based   & \yes & \no  & \yes & \yes & \no  & \yes & \pa  \\ 
Anthropic Claude Code   & \yes & \yes  & \yes & \no & \pa  & \no & \pa  \\ 
\rowcolor{cliffteal!44}
\textbf{\textsc{CliffCompaction}} & \yes & \yes & \yes & \yes & \yes & \yes & \no  \\
\bottomrule
\end{tabular}%
}
\end{table}

%% file: contents/method.tex
\label{sec:method}

\textsc{CliffCompaction} makes three design choices. First, it compacts when the context crosses a token threshold. Second, it retains compact verbatim excerpts---rather than summaries---while dropping token-intensive portions of the conversation. Third, at each new compaction event, it discards previous compactions and compacts only the post-compaction turns since the last event. This leads to a "cliff" where context length drops sharply to roughly the same level after each compaction event.

This design aims to optimize two main objectives: (1) making compaction itself practical and efficient to deploy; (2) maximizing verbatim, unaltered context for as long as possible. 

While the first objective simply avoids re-prefills, the second is more subtle. Using summaries has the problem that if {\it most} of the information is summarized correctly, the model assumes it has all the information and does not need to revisit past state, such as documentation or code files (Appendix~\ref{appendix:extended_analysis}, Figure~\ref{fig:extra-steps}). This leads to subtle but continuous context drift if the summary does not preserve the right information. We call this issue low {\it compaction precision}, where high compaction precision avoids omission or distortion of past information in the context. 

This often stands in trade-off with {\it compaction recall}: how much information can be retrieved over the course of many compactions.

\textsc{CliffCompaction} can be seen as having low compaction recall, since it throws away all direct information after two compactions. Summaries have high recall since they can retrieve information from many past compactions. On the other hand, \textsc{CliffCompaction} has high compaction precision, because it preserves user directives and model generations in full for two full compaction windows, while methods that use summaries have low compaction precision.

As such, the best view of \textsc{CliffCompaction} is as a practically deployable algorithm that optimizes compaction precision at the cost of degrading compaction recall.

\begin{figure}[ht]
  \centering
  \includegraphics[width=\linewidth]{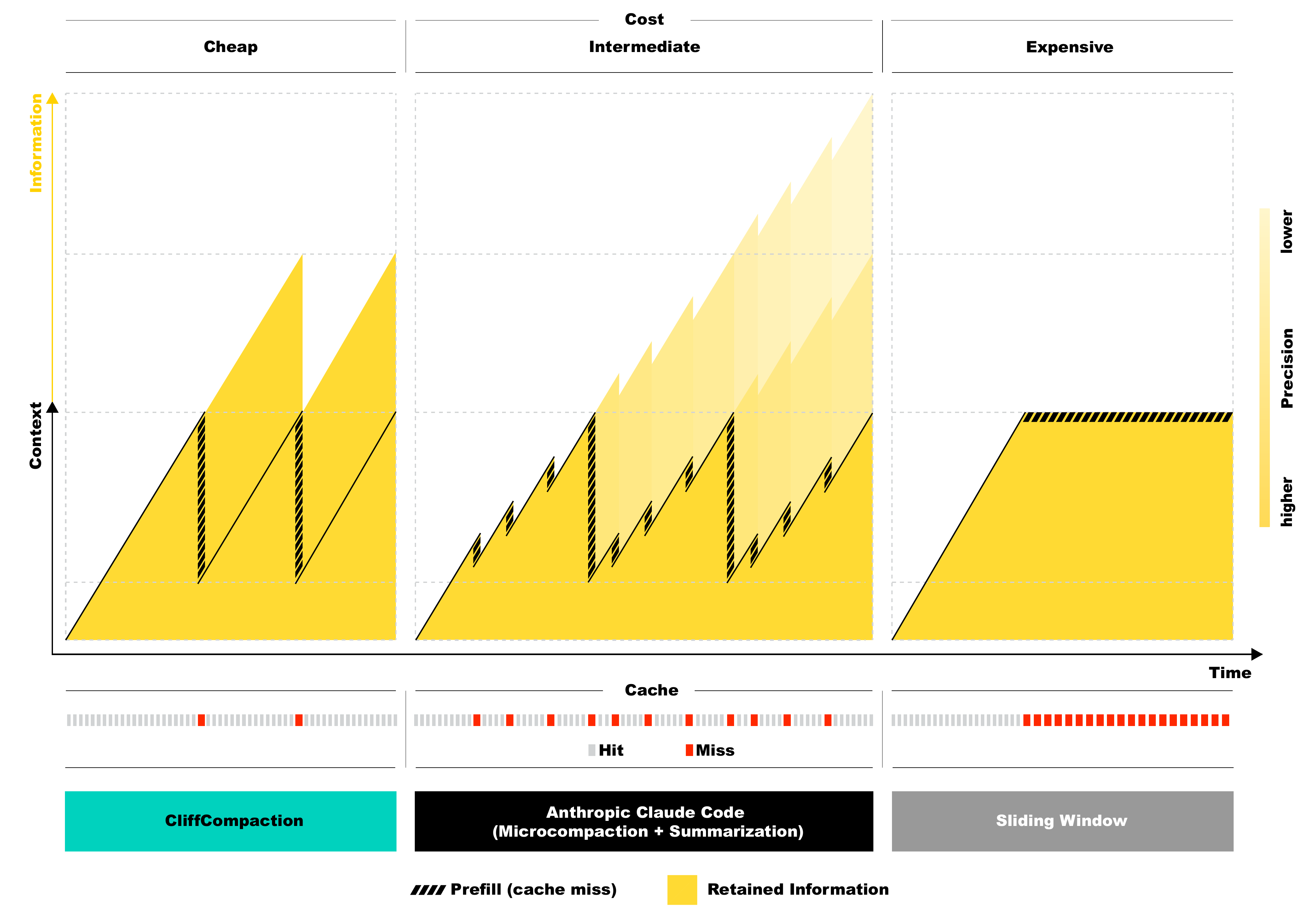}
  \caption{\textbf{Overview of \textsc{CliffCompaction}.} \textsc{CliffCompaction} strictly discards previous compactions and retains only fragments from the most recent window. The yellow area reflects recall (how much information stays retrievable in the context) and its shade reflects precision (fidelity of the retained information)---\textsc{CliffCompaction} favors precision over recall.}
  \label{fig:cliff}
  \vspace{-5pt}
\end{figure}



\subsection{\textsc{CliffCompaction} and KV-Cache}
\label{sec:cliff_and_cache}
On the frontend level, LLM APIs typically distinguish between uncached input tokens, output tokens, and cached input tokens. On the backend systems level, cached input tokens correspond to KV-cache reads. While cache reads are the least expensive on a per-token basis, they are by far the most costly operation for long-context sessions that are typical for agents. This is because every request bears the cost of all past tokens while autoregressive token generation incurs only the cost of a few additional tokens. On the backend level, this cost is related to large memory reads for KV-caches, which usually take up more GPU memory than the model weights and which are bandwidth-bound---this is particularly expensive due to the current shortage and price of HBM memory. 

Any modification to the context, including compaction and other context-management approaches, invalidates the existing KV-cache and requires re-prefilling the new context from the common prefix. API costs for uncached input tokens that require re-prefill are roughly 5 to 6 times as expensive as cached ones in the models we study (Table~\ref{tab:model-prices}) and as such incur very high costs. Consequently, the efficiency gains of frontend context management often come at the expense of additional backend re-prefills. Thus frequent context changes need to be avoided for a practical, cost-efficient algorithm.

\textsc{CliffCompaction} is designed to preserve cache efficiency by avoiding excessive re-prefills. Instead of actively managing the context throughout a trajectory, \textsc{CliffCompaction} leaves it unchanged and lets it grow naturally, compacting only upon exhausting a preset budget. Since the existing context are never modified between compactions, the cache remains valid across each entire growth segment and is invalidated only at the compaction points themselves, hence the smooth cliff-shaped profile in Figure~\ref{fig:cliff}. While each compaction event unavoidably resets the cache and requires a re-prefill, the associated overhead stays modest for two reasons: the compacted context is dramatically smaller than the original, and compaction occurs only occasionally, so each re-prefill is amortized over a long stretch of full cache reuse. 


\subsection{\textsc{CliffCompaction} and Tokens}

We unpack the agent's context on Terminal-Bench 2.0 with GLM 5.1 to identify where tokens and dollars are concentrated (Table~\ref{tab:cost-attribution}). Without compaction, tool results dominate the context at 56.0\% of tokens, followed by tool calls at 28.0\%, together accounting for 84\% of the total spent. This motivates \textsc{CliffCompaction} to compact tool-related content by dropping long tool results and reducing tool calls to their signatures. If the model wants to retrieve past information, it can issue a new tool call from the preserved signature, so any removed context remains recoverable.

\paragraph{Tool results.}
\textsc{CliffCompaction} applies a simple length-based rule to tool results. Those exceeding $500$ characters are dropped, while shorter ones are retained. Short results are often useful and cheap to keep, such as grep matches, exit codes, and concise script output. Long results are typically reads of large files and contain information that can be recovered on demand.

\paragraph{Tool calls.}
For tool calls, \textsc{CliffCompaction} reduces each call to a compact signature. Although available tools differ across scaffolds, it generally retains the tool name, the target file or path, and other essential arguments. Long contents embedded in tool calls, such as file-write calls that inline file contents, are removed, as the resulting file persists in the codebase and can be re-read whenever needed. By preserving tool calls verbatim, the agent can recover past context fully by making the same tool call.

\paragraph{Thoughts, system prompts, task description, and recent turns.} Agent thoughts (analyses, plans, and hypotheses) are not a major source of context bloat, but we still truncate them to $300$ characters. The system prompt and the first user message (the task description) are kept in full. We also keep the $K$ most recent turns unchanged (Algorithm~\ref{alg:cliff}).



\subsection{\textsc{Cliff} by \textsc{Cliff}}

Long trajectories may undergo multiple compactions, raising the question of how to manage previous compactions when a new one fires. \textsc{CliffCompaction} makes a trade-off where we compact past sessions in such a way to maximize compaction precision---how much information to preserve verbatim---while limiting degradations in compaction recall---how much information is preserved from past sessions.  

In more detail, the \textsc{CliffCompaction} algorithm works as follows: After the $(t{-}1)$-th compaction, the agent's context $S_t$ consists of the compacted history $C_{t-1}$ followed by the live session $L_t$ that grows on top of it, with $C_0 = \emptyset$. When $S_t$ exceeds the context window, the $t$-th compaction triggers:
\begin{align*}
S_t     &= C_{t-1} \oplus L_t, \\
C_t     &= \textsc{CliffCompaction}(L_t), \\
S_{t+1} &= C_t \oplus L_{t+1}.
\end{align*}

$C_{t-1}$ is discarded entirely rather than nested into $C_t$.

\paragraph{Faithfulness---maximizing compaction precision.} Common compaction schemes fold old summaries into new ones, so every cascading compaction is a summary of summaries. Compressing already partial content makes each compaction lossier than the last, and context quality decays over the trajectory. \textsc{CliffCompaction} prevents this type of regression by discarding the previous compacted history completely. Each compaction compresses only the latest live session $L_t$, keeping every cliff equally faithful to the turns it condenses and maintaining high compaction precision throughout the trajectory.

\paragraph{Residual propagation limits degradation of compaction recall.} While \textsc{CliffCompaction} discards all information before the previous compaction, it still carries information forward. Every turn of $L_t$ is generated with $C_{t-1}$ in view, so the agent's actions and reasoning, and therefore the block $C_t$ that compresses them, carry an implicit influence of the discarded context. This chain extends back to the start of the trajectory:
\[
C_t \leftarrow L_t \leftarrow C_{t-1} \leftarrow L_{t-1}
\leftarrow \cdots \leftarrow C_1 \leftarrow L_1.
\]
No information is ever carried forward directly, but residual
knowledge leaks from each session into the next through the agent's own behavior, softening the loss in compaction recall. Experimentally, this residual knowledge makes \textsc{CliffCompaction} effective at long-context sessions over millions of tokens even though a compaction window of 128k tokens is used.

As such, the two effects trade recall for precision. Only the most recent window survives each compaction, so the context is cliff-shaped not only in tokens but also in information (Figure~\ref{fig:cliff}). Algorithm~\ref{alg:cliff} summarizes \textsc{CliffCompaction}.

\begin{algorithm}[t]
\caption{\textsc{CliffCompaction}}
\label{alg:cliff}
\begin{algorithmic}[1]
\Function{CliffCompaction}{$\mathit{turns}$}
  \State $\mathit{old}, \mathit{recent} \gets \mathit{turns}[{:}{-}2K],\ \mathit{turns}[{-}2K{:}]$ \Comment{$K$ turn pairs}
  \State $\mathit{parts} \gets [\,]$
  \For{$m \in \mathit{old}$}
    \If{$m$ is a prior compaction}
      \State \textbf{skip} \Comment{discard, keep flat}
    \ElsIf{$m$ is \textsc{Assistant}}
      \State $\mathit{parts} \mathrel{+}= [\textsc{Truncate}(m.\mathit{thinking}, 300),\ \textsc{Signature}(m.\mathit{toolcall}, 150)]$
    \ElsIf{$m$ is \textsc{ToolResult} \textbf{and} $|m| \le 500$}
      \State $\mathit{parts} \mathrel{+}= [m]$
    \EndIf
  \EndFor
  \State \Return $\textsc{Join}(\mathit{parts}),\ \mathit{recent}$
\EndFunction
\Statex
\Function{Query}{$\mathit{messages}$}
  \State \textbf{try} \Return $\textsc{LLM}(\mathit{messages})$
  \Except{\textsc{ContextWindowExceeded}}
    \State $s, x \gets \mathit{messages}[0],\ \mathit{messages}[1]$
    \State $C, \mathit{recent} \gets \Call{CliffCompaction}{\mathit{messages}[2{:}]}$
    \State \Return $\textsc{LLM}([s, x, C]\ \|\ \mathit{recent})$
  \EndExcept
\EndFunction
\end{algorithmic}
\end{algorithm}

%% file: contents/mainresults.tex
\subsection{Experimental Setup}

\paragraph{Benchmarks, Scaffolds, and Models.}
We evaluate on SWE-bench Verified~\citep{jimenez2024swebench} and Terminal-Bench~\citep{merrill2026terminal}. For SWE-bench Verified, we use \texttt{mini-swe-agent}~\citep{yang2024sweagent}, which maintains an append-only conversation history without native conversation-level compaction, providing a clean baseline. We additionally use \texttt{OpenHands}~\citep{wang2025openhands} to test whether the effect transfers across harnesses, replacing its native condenser with \textsc{CliffCompaction}{} in the constrained-context settings. For Terminal-Bench 2.0, we use the benchmark's standard \texttt{Terminus-2} scaffold, substituting its built-in compaction with \textsc{CliffCompaction}{} while leaving all other behavior unchanged. We also evaluate \textsc{CliffCompaction}{} on Terminal-Bench 2.1 using \texttt{Claude Code}. We run on recent models from the Kimi (K2.5, K2.6)~\citep{team2026kimi, moonshot2026kimik26} and GLM (5, 5.1, 5 Turbo, 5.3 Flash, 4.7 Flash)~\citep{glm5team2026glm5vibecodingagentic, zai2026glm53flash, 5team2025glm45agenticreasoningcoding} families.

\paragraph{Compaction Thresholds and Integration.}
We compare full-context execution under each scaffold's default behavior against \textsc{CliffCompaction}{} with token thresholds $B \in \{45\text{K}, 32\text{K}, 16\text{K}, 8\text{K}\}$. For \texttt{mini-swe-agent}, \texttt{OpenHands} and \texttt{Terminus-2} we modify the scaffold and integrate \textsc{CliffCompaction}{} directly. \texttt{Claude Code} is closed-source, so we implement \textsc{CliffCompaction}{} as an API proxy that works with any scaffold. Both \textsc{CliffCompaction}{} and \texttt{Claude Code}'s native auto-compaction are configured to operate at a matched mean peak context of ${\sim}45$K.

\subsection{Results}
We first evaluate whether \textsc{CliffCompaction}{} preserves agent success rates when the available context is substantially constrained. Tables~\ref{tab:swebench-verified} and~\ref{tab:terminal-bench} compare runs using the model's full context window against runs with fixed compaction thresholds on SWE-bench Verified and Terminal-Bench. We focus here on task success and analyze the cost columns in Appendix~\ref{sec:cost}.

\begin{table}[!htbp]
  \centering
 \caption{Performance and cost on Terminal-Bench across context length budgets with
  \textsc{CliffCompaction}, evaluated with \texttt{Terminus-2} on Terminal-Bench 2.0 and
  \texttt{Claude Code} on Terminal-Bench 2.1. $^{\dagger}$No compaction is applied. \\}
  \label{tab:terminal-bench}
  \setlength{\tabcolsep}{4pt}
  \resizebox{\linewidth}{!}{%
  \begin{tabular}{l l cc cc cc cc}
    \toprule
    \multicolumn{10}{@{}c}{\texttt{Terminus-2}} \\
    \midrule
    & & \multicolumn{2}{c}{Full context$^{\dagger}$} & \multicolumn{2}{c}{32K}
    & \multicolumn{2}{c}{16K} & \multicolumn{2}{c}{8K} \\
    \cmidrule(lr){3-4} \cmidrule(lr){5-6} \cmidrule(lr){7-8} \cmidrule(lr){9-10}
    Model & Compaction & \% Resolved & Cost & \% Resolved & Cost
             & \% Resolved & Cost & \% Resolved & Cost \\
    \midrule
    \multirow{2}{*}{Kimi K2.6} & Summarization
                            & \multirow{2}{*}{$59.16_{\pm 3.41}$} & \multirow{2}{*}{\$0.40}
                            & $58.36_{\pm 2.57}$ & \$0.24
                            & $55.45_{\pm 0.64}$ & \$0.26
                            & $42.97_{\pm 4.41}$ & \$0.60 \\
    &  \textsc{CliffCompaction}
                            &                    &
                            &  $61.42_{\pm 4.25}$ &  \$0.24
                            &  $61.42_{\pm 1.30}$ &  \$0.19
                            &  $50.00_{\pm 2.40}$ &  \$0.25 \\
    \midrule
    GLM 5.1 &  \textsc{CliffCompaction} & $49.83_{\pm 3.42}$ & \$0.54
                            &  $53.20_{\pm 5.31}$ &  \$0.34
                            &  $54.33_{\pm 2.35}$ &  \$0.27
                            &  $44.93_{\pm 3.35}$ &  \$0.19 \\
    \midrule
    \addlinespace[2pt]
    \multicolumn{10}{@{}c}{\texttt{Claude Code}} \\
    \midrule
    & & \multicolumn{4}{c}{200K} & \multicolumn{4}{c}{45K} \\
    \cmidrule(lr){3-6} \cmidrule(lr){7-10}
    Model & Compaction & \multicolumn{2}{c}{\% Resolved} & \multicolumn{2}{c}{Cost}
             & \multicolumn{2}{c}{\% Resolved} & \multicolumn{2}{c}{Cost} \\
    \midrule
    \multirow{2}{*}{GLM 5.3 Flash}
                  & Summarization
                  & \multicolumn{2}{c}{$73.03_{\pm 3.04}$} & \multicolumn{2}{c}{\$0.21}
                  & \multicolumn{2}{c}{$70.97_{\pm 3.72}$} & \multicolumn{2}{c}{\$0.14} \\
    
                  & \textsc{CliffCompaction}
                  & \multicolumn{2}{c}{---} & \multicolumn{2}{c}{---}
                  & \multicolumn{2}{c}{$76.69_{\pm 1.41}$} & \multicolumn{2}{c}{\$0.16} \\

    \bottomrule
  \end{tabular}%
  }
\end{table}

\paragraph{Terminal-Bench.}
On Terminal-Bench 2.0, \textsc{CliffCompaction}{} matches or improves upon the full-context setting at moderate thresholds. For Kimi K2.6, both the 32K and 16K settings achieve $61.42\%$, exceeding the full-context baseline of $59.16\%$ by $2.26$ percentage points. At 16K, replacing \texttt{Terminus-2}'s native LLM-based summarization with \textsc{CliffCompaction}{} increases success from $55.45\%$ to $61.42\%$. GLM 5.1 shows a similar benefit, improving from $49.83\%$ at full context to $53.20\%$ at 32K and $54.33\%$ at 16K. These gains suggest that removing stale context can sometimes improve agent performance rather than simply reducing memory pressure. As in SWE-bench Verified, the 8K setting is substantially more constrained and shows lower success, but the degradation remains gradual rather than catastrophic.

On Terminal-Bench 2.1 we evaluate GLM 5.3 Flash under \texttt{Claude Code}. At a matched ${\sim}45$K mean peak context, \textsc{CliffCompaction}{} reaches $76.69\%$, exceeding both \texttt{Claude Code}'s auto-compaction at the same budget ($70.97\%$) and its default 200K configuration ($73.03\%$), and doing so with markedly lower seed variance. This also highlights that \textsc{CliffCompaction} is deliberately scaffold-agnostic: it has no knowledge of \texttt{Claude Code}'s tool schema and reduces every tool call to a generic name-and-arguments signature. That \textsc{CliffCompaction}{} still outperforms a scaffold-native summarizer under these conditions indicates the method does not depend on scaffold-specific structure, and can be deployed against a closed agent without modification.

\input{tables/swe-bench}

\paragraph{SWE-bench Verified.}
On SWE-bench Verified, \textsc{CliffCompaction}{} preserves most of the full-context performance at moderate thresholds. For Kimi K2.6, reducing the compaction threshold to 32K changes the resolution rate only slightly, from $73.87\%$ to $73.27\%$, and the 16K setting remains within $2.0$ percentage points of the full-context baseline. The GLM models show a similar trend, with GLM 5.1 and GLM-5 Turbo remaining within roughly two percentage points of their full-context baselines at 16K. The 8K setting is more aggressive and produces larger drops across models, suggesting it is best viewed as a stress test rather than the default operating point.


%% file: tables/swe-bench.tex
\begin{table}[!htbp]
  \centering
  \caption{Performance and cost on SWE-bench Verified across context length budgets,
  evaluated with \texttt{mini-swe-agent} and \texttt{OpenHands}. Cost is average per-instance USD. $^{\dagger}$No compaction is applied. $^{\ddagger}$\texttt{OpenHands}' native condenser is active. \\}
  \label{tab:swebench-verified}
  \footnotesize
  \setlength{\tabcolsep}{4pt}
  \begin{tabular}{l cc cc cc cc}
    \toprule
    \multicolumn{9}{@{}c}{\texttt{mini-swe-agent}} \\
    \midrule
    & \multicolumn{2}{c}{Full context$^{\dagger}$} & \multicolumn{2}{c}{32K}
    & \multicolumn{2}{c}{16K} & \multicolumn{2}{c}{8K} \\
    \cmidrule(lr){2-3} \cmidrule(lr){4-5} \cmidrule(lr){6-7} \cmidrule(lr){8-9}
    Model & \% Resolved & Cost & \% Resolved & Cost
          & \% Resolved & Cost & \% Resolved & Cost \\
    \midrule
    Kimi K2.6     & $73.87_{\pm 0.31}$      & \$0.19
                  & $73.27_{\pm 0.12}$      & \$0.18
                  & $71.87_{\pm 1.31}$      & \$0.17
                  & $67.60_{\pm 1.06}$      & \$0.18 \\
    Kimi K2.5     & $70.87_{\pm 1.01}$      & \$0.12
                  & $70.98_{\pm 0.98}$      & \$0.11
                  & $69.53_{\pm 1.33}$      & \$0.08
                  & $64.93_{\pm 1.33}$      & \$0.10 \\
    GLM 5.1       & $71.40_{\pm 0.72}$      & \$0.25
                  & $68.80_{\pm 1.59}$      & \$0.20
                  & $69.33_{\pm 1.10}$      & \$0.15
                  & $65.13_{\pm 0.76}$      & \$0.13 \\
    GLM 5 Turbo   & $69.80_{\pm 1.06}$      & \$0.19
                  & $70.53_{\pm 0.42}$      & \$0.16
                  & $67.53_{\pm 1.14}$      & \$0.13
                  & $63.27_{\pm 1.22}$      & \$0.11 \\
    GLM 5         & $69.80_{\pm 1.20}$      & \$0.26
                  & $68.87_{\pm 0.81}$     & \$0.23
                  & ---      & --- 
                  & ---                     & ---    \\
    GLM 4.7 Flash & $43.20_{\pm 1.27}$      & ---
                  & $43.60_{\pm 0.99}$      & ---
                  & $41.20_{\pm 0.42}$      & ---
                  & $30.70_{\pm 2.97}$      & ---    \\
    \midrule
    \addlinespace[2pt]
    \multicolumn{9}{@{}c}{\texttt{OpenHands}} \\
    \midrule
    & \multicolumn{4}{c}{Full context$^{\ddagger}$} & \multicolumn{4}{c}{32K} \\
    \cmidrule(lr){2-5} \cmidrule(lr){6-9} 
    Model & \multicolumn{2}{c}{\% Resolved} & \multicolumn{2}{c}{Cost}
          & \multicolumn{2}{c}{\% Resolved} & \multicolumn{2}{c}{Cost} \\
    \midrule
    Kimi K2.6     & \multicolumn{2}{c}{$71.30_{\pm 0.99}$} & \multicolumn{2}{c}{\$0.57} 
                  & \multicolumn{2}{c}{$70.20_{\pm 0.57}$} & \multicolumn{2}{c}{\$0.42} \\
    GLM 5.1       & \multicolumn{2}{c}{$71.93_{\pm 1.21}$} & \multicolumn{2}{c}{\$1.01} 
                  & \multicolumn{2}{c}{$72.20_{\pm 0.69}$} & \multicolumn{2}{c}{\$0.69} \\
    \bottomrule
  \end{tabular}
\end{table}

%% file: contents/tts.tex



Test-time scaling improves agent performance by performing multiple rollouts for a task and then using an algorithm, model, or heuristic to choose the best solution, but it has remained mostly academic due to its high cost. To put the cost-performance trade-off into perspective: without \textsc{CliffCompaction} three rollouts of Kimi K2.6 on all Terminal-Bench instances buy 4.8 points of performance gain over a single run at three times its cost (\$91.65)---a price that exceeds a single run of a proprietary model that scores 5.5 points higher at a cost of \$65, which shows the impractical nature of naive test-time scaling. 

\textsc{CliffCompaction} inverts this trade-off. Compaction reduces per-rollout cost enough that multiple rollouts can fit within the budget of a single frontier-model run. With \textsc{CliffCompaction}, three Kimi K2.6 rollouts cost \$58.01 for the entirety of Terminal-Bench, less than one run of GPT 5.3 Codex, while matching the strongest proprietary model, Anthropic's Opus 4.7 (see Table~\ref{tab:terminalbench-scaling}).

\vspace{-3pt}
\subsection{Experimental Setup} 
\vspace{-2pt}

We scale \texttt{Terminus-2} on Terminal-Bench (Table~\ref{tab:terminalbench-scaling}) and \texttt{mini-swe-agent} on SWE-bench Verified (Table~\ref{tab:swebench-scaling}) $k$ times under each \textsc{CliffCompaction} context threshold, for both Kimi K2.6 and GLM 5.1. We report \textit{pass@1}, the mean resolution rate across individual runs (without selection), and \textit{Oracle} (\textit{pass@$k$}), the fraction of tasks solved by at least one of the $k$ runs, effectively picking the best trajectory among all rollouts, which serves as an upper bound on achievable performance.

Since oracle selection is unavailable at deployment time, we additionally report \textit{Practical}, the resolution rate achieved by a learned selector. For each candidate trajectory $T$, we extract a feature vector $\phi(T) = [f_1(T), \dots, f_d(T)] \in \mathbb{R}^d$ that summarizes its execution behavior (e.g., the number of steps and tool calls) and characteristics of the produced solution (e.g., its length and structure). We further find line overlap to be a strong statistical feature that substantially boosts accuracy for trajectory selection. This aligns with \citet{shen2026sera}, who show that the line overlap of two rollouts correlates strongly with unit-test resolution. Following this, we generalize line overlap to a broader set of within-group agreement features that capture how much a candidate agrees with the others for the same task. We call the resulting selector \textsc{Soft Group Verification} (SGV).

A lightweight LightGBM classifier $s(\cdot)$ takes these features as inputs and assigns each candidate a score $s(\phi(T))$, and we select $\hat{T} = \arg\max_T s(\phi(T))$ as the final solution. The scorer is trained on rollouts from a \emph{different} model: the selector used on Kimi K2.6 is trained only on GLM 5.1's, and vice versa. We find no consistent difference between selectors trained on the same versus a different model. We additional test for test leakage and find SGV to not leak any information through summary statistics of the trajectory. Details are in Appendix~\ref{appendix:tts}.

\vspace{-3pt}
\subsection{Results}
\vspace{-2pt}

\begin{table}[!htbp]
  \centering
  \caption{Test-time scaling on Terminal-Bench 2.0, ordered by accuracy. With \textsc{CliffCompaction} (abbreviated \textsc{Cliff} here for space), three rollouts of Kimi K2.6 cost less than a single GPT 5.3 Codex run while surpassing every proprietary baseline evaluated on the same \texttt{Terminus-2} harness. \emph{Practical} is the resolution rate of our learned selector, \textsc{Soft Group Verification} (SGV); at $k{=}1$ it is plain pass@1. \emph{Oracle} is the pass@$k$ ceiling. \emph{Context} is the compaction budget for \textsc{Cliff} rows and the model's full window otherwise. Teal rows are Pareto-optimal in accuracy--cost among our scaled ($k{>}1$) configurations; purple rows are proprietary baselines.}
  \label{tab:terminalbench-scaling}
  \footnotesize
  \setlength{\tabcolsep}{5pt}
  \resizebox{\textwidth}{!}{%
  \begin{tabular}{l l l c r c c c}
    \toprule
    Model & Scaffold & Context & $k$ & Cost (USD) & Oracle & Practical & Gain\,@\,Cost\textsuperscript{*} \\
    \midrule
    \rowcolor{citepurple!12}
    GPT 5.5 + LLM-as-a-Verifier\textsuperscript{\S} & \texttt{Capy} & 1M & 5 & ---              & 92.1\textsuperscript{\S} & 86.5\textsuperscript{\S} & +3.4\,@\,5.0$\times$ \\
    \rowcolor{citepurple!12}
    GPT 5.5\textsuperscript{\S}                     & \texttt{Capy} & 1M & 1 & ---              & ---  & 83.1\textsuperscript{\S} & --- \\
    \rowcolor{cliffteal!44}
    Kimi K2.6 \textsc{+ Cliff + SGV} & \texttt{Terminus-2} & 16K  & 3 & \$58.01          & 74.2 & 69.7 & +10.5\,@\,1.9$\times$ \\
    \rowcolor{citepurple!12}
    Opus 4.7\textsuperscript{$\ddagger$}            & ---  & 1M & 1 & ---              & ---  & 69.4 & --- \\
    \rowcolor{cliffteal!44}
    Kimi K2.6 \textsc{+ Cliff + SGV} & \texttt{Terminus-2} & 16K  & 2 & \$38.67          & 70.4 & 65.9 & +6.7\,@\,1.3$\times$ \\
    Kimi K2.6 \textsc{+ Cliff + SGV} & \texttt{Terminus-2} & 32K  & 3 & \$65.33          & 73.0 & 65.2 & +6.0\,@\,2.1$\times$ \\
    \rowcolor{citepurple!12}
    GPT 5.3 Codex              & \texttt{Terminus-2} & 272K & 1 & \$64.63\textsuperscript{$\dagger$} & --- & 64.7 & --- \\
    Kimi K2.6 \textsc{+ SGV}   & \texttt{Terminus-2} & 256K & 3 & \$91.65          & 70.8 & 64.0 & +4.8\,@\,3.0$\times$ \\
    \rowcolor{citepurple!12}
    Opus 4.6                   & \texttt{Terminus-2} & 1M & 1 & \$44.53\textsuperscript{$\dagger$} & --- & 62.9 & --- \\
    GLM 5.1 \textsc{+ Cliff + SGV}   & \texttt{Terminus-2} & 32K  & 3 & \$95.34          & 68.5 & 60.7 & +10.9\,@\,1.8$\times$ \\
    GLM 5.1 \textsc{+ Cliff + SGV}   & \texttt{Terminus-2} & 16K  & 3 & \$87.25          & 68.5 & 59.6 & +9.8\,@\,1.7$\times$ \\
    Kimi K2.6                  & \texttt{Terminus-2} & 256K & 1 & \$30.55          & --- & 59.2 & --- \\
    GLM 5.1 \textsc{+ SGV}     & \texttt{Terminus-2} & 200K & 3 & \$158.25         & 65.2 & 55.1 & +5.3\,@\,3.0$\times$ \\
    GLM 5.1                    & \texttt{Terminus-2} & 200K & 1 & \$52.75          & --- & 49.8 & --- \\
    \bottomrule
  \end{tabular}%
  }

  \vspace{3pt}
  \begin{minipage}{\textwidth}
    \footnotesize
    \setlength{\parindent}{0pt}
    \textsuperscript{*}\,Accuracy gain and cost multiplier relative to the same model's full-context single run ($k{=}1$).\par
    \textsuperscript{$\dagger$}\,Costs from the Terminal-Bench leaderboard (\texttt{Terminus-2}): Opus 4.6 uses the reported cost; GPT 5.3 Codex is computed from its token counts.\par
    \textsuperscript{$\ddagger$}\,Score from the Claude Opus 4.7 system card.\par
    \textsuperscript{\S}\,From \citet{kwok2026llmasaverifiergeneralpurposeverificationframework}; Gemini-2.5-Flash as the verifier.\par
  \end{minipage}
\end{table}

\textsc{CliffCompaction} makes test-time scaling cost-effective where naive scaling is not. At the same three-rollout budget, uncompacted Kimi K2.6 gains 4.8 points over a single run at $3.0\times$ its cost, while \textsc{CliffCompaction} at a 16K context limit gains 10.5 points at $1.9\times$ (Table~\ref{tab:terminalbench-scaling}). This configuration reaches 69.7\%, matching the strongest proprietary model we test (Opus 4.7, 69.4\%) and exceeding GPT 5.3 Codex (64.7\%) and Opus 4.6 (62.9\%), at a total cost of \$58.01, less than a single GPT 5.3 Codex run. It also dominates the uncompacted three-rollout baseline: 5.7 points more accurate at 37\% lower cost.

The gains persist under a more conservative test-time budget. With only two rollouts, \textsc{CliffCompaction} reaches 65.9\% at \$38.67, a $1.3\times$ cost multiplier, cheaper than a single run of either GPT 5.3 Codex or Opus 4.6. This already exceeds the uncompacted three-rollout baseline at 42\% of its cost.

The same pattern holds for GLM 5.1: \textsc{CliffCompaction} raises the practical resolution rate from 55.1\% to 60.7\% while reducing scaled-inference cost by 40\% (\$95.34 vs.\ \$158.25), narrowing much of the gap to the proprietary models.

To further illustrate the cost-effectiveness of \textsc{CliffCompaction} and \textsc{SGV}, we focus on \emph{Gain@Cost} and compare against another selector. Using a strong base model and a separate Gemini-2.5-Flash verifier to rank five rollouts, LLM-as-a-Verifier converts a $5.0\times$ cost increase into $+3.4$ points---the lowest gain-per-cost in the table. \textsc{CliffCompaction} with \textsc{SGV} turns a $1.9\times$ cost increase into $+10.5$ points, and $+6.7$ at only $1.3\times$. GLM 5.1 shows the same pattern ($+10.9$ at $1.8\times$).

%% file: contents/continuallearning.tex
\label{sec:kernelbench}



Having established that \textsc{CliffCompaction} maintains or improves performance across benchmarks, scaffolds, and scales, we next investigate whether it can support continual learning over long horizons and many compactions. We evaluate on KernelBench Level~3~\citep{ouyang2025kernelbenchllmswriteefficient}, where a single trajectory can exceed one million tokens which is $15$--$70\times$ the median SWE-bench or Terminal-Bench trajectory.

\subsection{Experimental Setup}

We use \texttt{OpenHands} as the scaffold and let the agent freely optimize GPU kernels until reaching a stopping condition, either exhausting the step budget or, in the non-compaction setting, reaching the model's maximum context window. We report \textit{Speedup}, the geometric mean speedup against the PyTorch Eager baseline, with per-problem speedups clamped to a maximum of $10\times$ and problems without a correct kernel scored as $0.1$, following the evaluation protocol of \citet{du2026adaexplore}; \textit{Acc.}, the percentage of problems for which the agent produces a correct kernel; and $>p$, the percentage of problems achieving a speedup greater than $p$. Each configuration adopts the kernel language and correctness tolerance of the specialized baseline it is positioned against: our CUDA configurations must satisfy $\texttt{atol} = \texttt{rtol} = \texttt{1e-2}$, matching \citet{dai2026cuda}, and our Triton configuration must satisfy $\texttt{atol} = \texttt{rtol} = \texttt{5e-2}$, matching \citet{du2026adaexplore}.

We run Kimi~K2.7 writing CUDA kernels and GPT-5-mini writing Triton kernels, both profiled on an L40S GPU. We additionally run \textsc{CliffCompaction} with Kimi~K2.6 on an RTX~Pro~6000 Blackwell GPU. Additional experimental details are provided in Appendix~\ref{appendix:kernelbench}.

\begin{table}[!htbp]
  \centering
  \caption{KernelBench Level~3 results. Top: specialized kernel-optimization agents ($^\dagger$as reported in the original papers; $^\ddagger$CUDA-Agent reports the geometric mean over correct solutions only). Middle: \textsc{CliffCompaction} on L40S; L40S/A6000 are essentially the same hardware, differing only in ECC support. Bottom: \textsc{CliffCompaction} on RTX Pro 6000 Blackwell. We report both steps and the number of kernels generated, since search-based methods propose one kernel candidate per step whereas agentic setups spend some steps exploring without writing kernel code.}
  \label{tab:kernel-bench}
  \footnotesize
  \setlength{\tabcolsep}{3pt}
  \resizebox{\textwidth}{!}{%
  \begin{tabular}{l ll cccc ccc}
    \toprule
    Method & Model & Scaffold & Speedup & Acc. & $>1.2\times$ & $>2\times$ & Steps & Kernels & Hardware \\
    \midrule
    DR.Kernel$^{\dagger}$ & Qwen3 14B (RL) & custom & $0.97\times$ & 84\% & 16\% & 8\% & --- & 56 & L40S/A6000 \\
    Iterative Refinement$^{\dagger}$ & GPT-5-mini & custom & $1.31\times$ & 100\% & 24\% & 12\% & --- & 50 & L40S/A6000 \\
    OpenEvolve+Memory$^\dagger$ & GPT-5-mini & custom & $1.47\times$ & 100\% & 28\% & 10\% & --- & 50 & L40S/A6000 \\
    AdaExplore$^\dagger$ & GPT-5-mini & custom & $1.55\times$ & 100\% & 28\% & 16\% & --- & 50 & L40S/A6000 \\
    AdaExplore$^\dagger$ & GPT-5-mini & custom & $1.78\times$ & 100\% & 36\% & 22\% & --- & 200 & L40S/A6000 \\
    CUDA-Agent$^{\dagger\ddagger}$ & Seed 1.6 (RL) & OpenHands & $1.80\times$ & 94\% & --- & --- & 200 & --- & H20 \\
    \midrule
    \rowcolor{cliffteal!44}
    \textsc{CliffCompaction} (128K) & GPT-5-mini & OpenHands & $1.57\times$ & 96\% & 60\% & 36\% & 50 & 15 & L40S/A6000 \\
    \rowcolor{cliffteal!44}
    \textsc{CliffCompaction} (128K) & GPT-5-mini & OpenHands & $2.09\times$ & 100\% & 78\% & 52\% & 200 & 82 & L40S/A6000 \\
    \rowcolor{cliffteal!44}
    \textsc{CliffCompaction} (128K) & GPT-5-mini & OpenHands & $2.21\times$ & 100\% & 82\% & 54\% & 400 & 166 & L40S/A6000 \\
    No compaction (256K) & Kimi K2.7 & OpenHands & $1.30\times$ & 86\% & 66\% & 50\% & 200 & --- & L40S/A6000 \\
    \rowcolor{cliffteal!44}
    \textsc{CliffCompaction} (128K) & Kimi K2.7 & OpenHands & $2.23\times$ & 94\% & 88\% & 54\% & 200 & --- & L40S/A6000 \\
    \rowcolor{cliffteal!44}
    \textsc{CliffCompaction} (128K) & Kimi K2.7 & OpenHands & $3.58\times$ & 96\% & 92\% & 86\% & 400 & --- & L40S/A6000 \\
    \midrule
    No compaction (256K) & Kimi K2.6 & OpenHands & $1.51\times$ & 96\% & 72\% & 32\% & 200 & --- & RTX 6000 \\
    \rowcolor{cliffteal!44}
    \textsc{CliffCompaction} (200K) & Kimi K2.6 & OpenHands & $1.85\times$ & 96\% & 78\% & 40\% & 200 & --- & RTX 6000 \\
    \rowcolor{cliffteal!44}
    \textsc{CliffCompaction} (128K) & Kimi K2.6 & OpenHands & $1.87\times$ & 96\% & 78\% & 46\% & 200 & --- & RTX 6000 \\
    \rowcolor{cliffteal!44}
    \textsc{CliffCompaction} (128K) & Kimi K2.6 & OpenHands & $2.48\times$ & 98\% & 90\% & 64\% & 400 & --- & RTX 6000 \\
    \bottomrule
  \end{tabular}%
  }
\end{table}

\begin{figure}[t]
  \centering
  \begin{minipage}{0.97\textwidth}
    \begin{subfigure}[t]{0.49\textwidth}
      \centering
      \includegraphics[width=\linewidth]{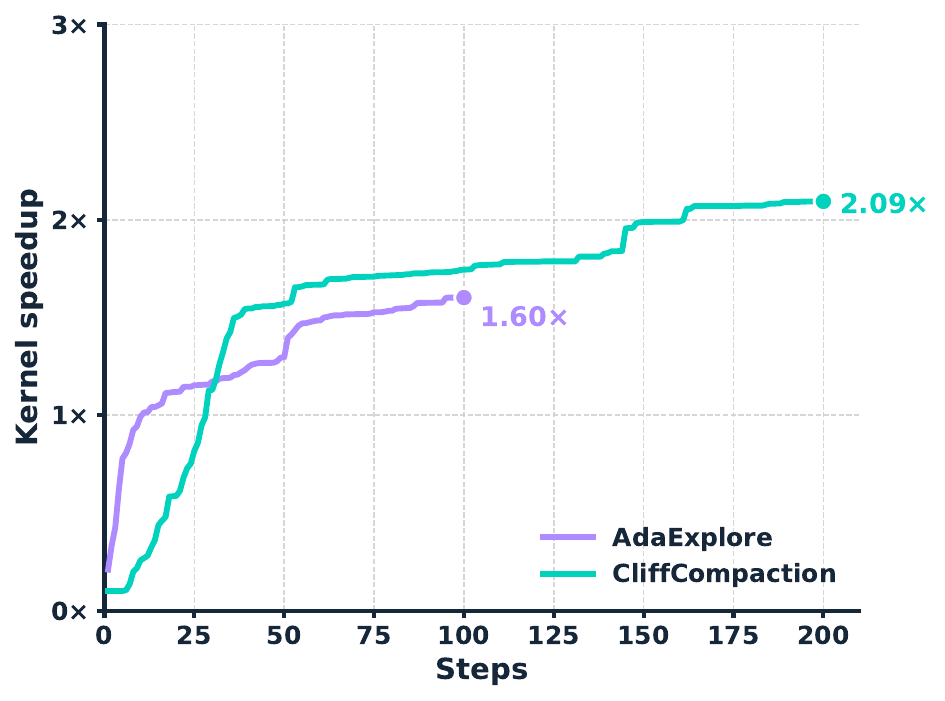}
      \caption{\textsc{CliffCompaction} vs.\ AdaExplore.}
      \label{fig:kb-continual}
    \end{subfigure}
    \hfill
    \begin{subfigure}[t]{0.49\textwidth}
      \centering
      \includegraphics[width=\linewidth]{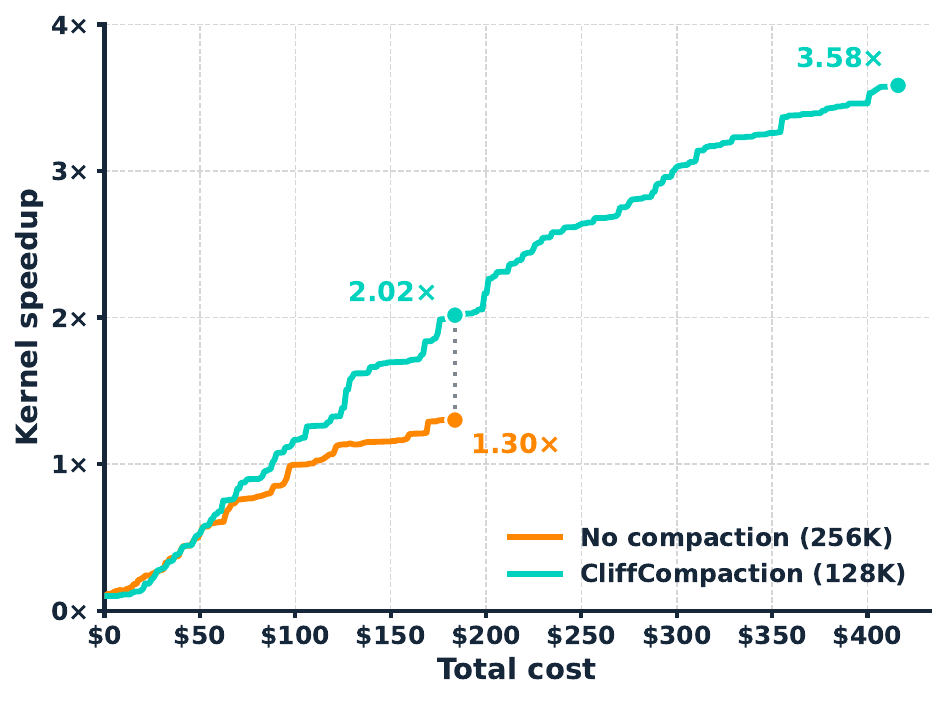}
      \caption{\textsc{CliffCompaction} vs.\ full context.}
      \label{fig:kb-cost}
    \end{subfigure}
  \end{minipage}
  \caption{Best-so-far kernel speedup on KernelBench Level~3, by steps and by cost.}
  \label{fig:kb-curves}
\end{figure}

\subsection{Results}

\paragraph{\textsc{CliffCompaction} sustains improvement beyond the context limit.} Without compaction, the agent exhausts the model's 256K context window well before the 200-step budget---98\% of runs on L40S and 76\% on RTX~Pro~6000 terminate early, at a median of only 99 and 122 steps---capping speedup at $1.30\times$ and $1.51\times$ (Table~\ref{tab:kernel-bench}). By letting the agent keep optimizing past this limit, \textsc{CliffCompaction} lifts speedup to $2.23\times$ and $1.87\times$ at 200 steps, and performance consistently rises when the budget is extended to 400 steps, reaching $3.58\times$ on L40S and $2.48\times$ on RTX~Pro~6000.

The gains are also broadly distributed across problems. On L40S, the fraction of kernels exceeding $2\times$ speedup climbs from 50\% without compaction to 86\% at 400 steps, and the fraction exceeding $1.2\times$ rises from 66\% to 92\%. On RTX~Pro~6000, $>\!2\times$ doubles from 32\% to 64\% while accuracy climbs to 98\%. Both the 200K and 128K compaction thresholds outperform the uncompacted 256K baseline, so the gains are robust to the threshold.

\paragraph{\textsc{CliffCompaction}'s bounded context is cheaper to scale.} Section~\ref{sec:tts} scaled test-time compute in parallel, here we apply the same idea along a sequential axis. With a smaller context window, \textsc{CliffCompaction} buys more optimization per dollar, so the same budget goes further. In Figure~\ref{fig:kb-cost}, at equal total cost \textsc{CliffCompaction} reaches $2.02\times$ against $1.30\times$ for the full-context run.

\paragraph{\textsc{CliffCompaction} outperforms systems purpose-built for kernel optimization.} At 200 agent steps it reaches $2.09\times$ against $1.78\times$ for AdaExplore on GPT-5-mini, and $2.23\times$ against $1.80\times$ for CUDA-Agent. Coverage-wise, at 200 steps \textsc{CliffCompaction} clears $1.2\times$ on 78\% of problems and $2\times$ on 52\%, more than double AdaExplore's 36\% and 22\%, spanning 82\% and 54\% at 400 steps. We note that the action unit differs between strategies, the comparison favors \textsc{CliffCompaction} further in candidates, where it achieves a higher speedup with fewer kernels proposed (Appendix~\ref{appendix:kernels_steps}).

Figure~\ref{fig:kb-continual} shows how speedup evolves for both. \textsc{CliffCompaction} begins well behind, because it must explore from scratch, whereas AdaExplore's learned skill memory starts strong. It catches up and overtakes at around step 31, remaining ahead through the end of AdaExplore's budget and continuing to improve well beyond it. In Table~\ref{tab:kernel-bench}, scaling from 50 to 200 steps yields $+0.64\times$ for \textsc{CliffCompaction}, compared with only $+0.23\times$ for AdaExplore. Given only compaction, a generic coding agent sustains productive optimization long after a purpose-built search has plateaued.

%% file: contents/compactioncomparison.tex
\label{sec:compaction_comparison}

\begin{table}[!htbp]
  \centering
  \caption{Performance and cost of compaction methods across benchmarks. SWE-bench Verified (16K) uses \texttt{mini-swe-agent}. $\Delta$ is change in cost vs.\ full-context (\$0.24 Kimi, \$0.15 GLM). KernelBench Level~3 uses \texttt{OpenHands} (128K) and a 400-step budget.}
  \label{tab:compaction-cost}
  \footnotesize
  \setlength{\tabcolsep}{4pt}
  \resizebox{\textwidth}{!}{%
  \begin{tabular}{l ccc ccc ccc}
    \toprule
    & \multicolumn{6}{c}{SWE-bench Verified} & \multicolumn{3}{c}{KernelBench L3} \\
    \cmidrule(lr){2-7} \cmidrule(lr){8-10}
    & \multicolumn{3}{c}{Kimi K2.7} & \multicolumn{3}{c}{GLM 5.2} & \multicolumn{3}{c}{Kimi K2.7} \\
    \cmidrule(lr){2-4} \cmidrule(lr){5-7} \cmidrule(lr){8-10}
    Method & \% Resolved & Cost & $\Delta$ & \% Resolved & Cost & $\Delta$ & Speedup & Kernels & Cost \\
    \midrule
    Sliding window          & $72.80_{\pm 2.43}$ & \$0.26 & $+9\%$  & $72.87_{\pm 0.99}$ & \$0.16 & $+7\%$  & $2.86\times$ & 118 & \$21.52 \\
    Summarization           & $70.27_{\pm 0.95}$ & \$0.18 & $-23\%$ & $70.27_{\pm 1.01}$ & \$0.12 & $-23\%$ & $3.47\times$ & 136 & \$8.10 \\
    \quad + Microcompaction   & $71.00_{\pm 0.69}$ & \$0.20 & $-17\%$ & $72.73_{\pm 2.08}$ & \$0.12 & $-23\%$ & $3.33\times$ & 129 & \$12.84 \\
    \midrule
    \rowcolor{cliffteal!44}
    \textsc{CliffCompaction} & $71.33_{\pm 0.50}$ & \$0.19 & $-21\%$ & $72.60_{\pm 1.22}$ & \$0.12 & $-23\%$ & $3.58\times$ & 116 & \$8.32 \\
    \bottomrule
  \end{tabular}%
  }
\end{table}

\paragraph{Setup.} We compare \textsc{CliffCompaction} against alternative context-management strategies (Table~\ref{tab:compaction-cost}). (1) \textbf{Sliding window} drops the oldest turns, keeping a fixed system/task prefix plus the most recent turns that fit within the context budget. (2) \textbf{Summarization} replaces the older turns with an LLM-generated summary, using Claude~Code's summarization prompt. (3) \textbf{Summarization + Microcompaction} is a two-tier pipeline that reimplements Claude Code's scheme: (i) \emph{microcompaction} clears the contents of old tool observations while preserving the corresponding tool calls and the $5$ most recent observations verbatim, adapting Anthropic's \texttt{clear\_tool\_uses} context-editing strategy~\citep{anthropic2025contextediting}; (ii) when microcompaction is insufficient, an LLM \emph{summarizer} rewrites the older turns into a structured summary using Claude~Code's summarization prompt. Microcompaction fires whenever at least $\tau$ tokens of reclaimable observation content have accumulated, with $\tau{=}20{,}000$ at a $128$K context budget and $\tau{=}4{,}000$ at $16$K. We adopt this configuration because it performs well on both quality and cost in our SWE-bench runs, making it a strong baseline.

\paragraph{\textsc{CliffCompaction} is the only method that stays competitive on quality while remaining cheap on every benchmark.} On SWE-bench Verified, where trajectories are short, all four methods fall within $2.6$ points of one another. Sliding window scores highest on both models but only marginally, and it is the only method that makes the agent \emph{more} expensive than running with full context ($+9\%$ and $+7\%$). Summarization is the weakest on quality for both models, echoing the pattern on Terminal-Bench (Table~\ref{tab:terminal-bench}). On KernelBench Level~3, with much longer trajectories, the methods separate more clearly: \textsc{CliffCompaction} reaches $3.58\times$ at \$8.32 per problem with the fewest kernel candidates (Appendix~\ref{appendix:kernels_steps}), vs.\ $3.47\times$ at \$8.10 for summarization, $3.33\times$ at \$12.84 with microcompaction, and $2.86\times$ at \$21.52 for sliding window. Across all three benchmarks, \textsc{CliffCompaction} is the only technique that both maintains strong performance and remains inexpensive.

%% file: contents/relatedwork.tex
\paragraph{Context management.}
Context management has long been an important research direction for long-horizon coding agents. Leading agent scaffolds manage growing trajectories by discarding earlier observations, implementing observation-level sliding-window compaction~\citep{yang2024sweagent, wang2025openhands}. Although simple, naively sliding the context can invalidate the KV cache and incur substantial re-prefill overhead. Another common approach uses LLMs to summarize prior context~\citep{wang2025openhands, kang2025acon, verma2026activecontextcompressionautonomous, li2026deepagent, wang2025recursively, li2026selfcompactinglanguagemodelagents}. Repeated summarization, however, can become increasingly lossy across compaction rounds, as details are progressively distorted or omitted through chains of summaries. A separate line of work develops trained context-compression mechanisms~\citep{sun2026scaling, kang2025acon, 10.1109/ASE63991.2025.00020, pan-etal-2024-llmlingua, jiang2026pabuprogressawarebeliefupdate}. While effective, these methods introduce additional training complexity and may not transfer readily across models. External-memory and retrieval-based approaches~\citep{packer2024memgptllmsoperatingsystems, wang2026memexrlscalinglonghorizonllm} preserve access to information outside the active context, but are not drop-in solutions---they require retrieval infrastructure and remain vulnerable to retrieval failures. Similarly, sub-agent-based approaches~\citep{gandhi2026recursiveagentoptimization, NEURIPS2024_ee71a4b1} introduce orchestration complexity and may require specialized prompting or training to determine when and how auxiliary agents should be invoked. Structure-based approaches~\citep{semenov2026compactionstructuredcontexteviction} use deterministic, rule-based policies to evict trajectory content, but require scaffold-specific instrumentation and are cache-inefficient.

\paragraph{Test-time scaling for code agents.}
Several recent works improve coding agent performance by spending additional inference-time compute on each instance. \citet{li2025s} introduce a hybrid framework that combines parallel sampling with sequential refinement and uses execution-grounded test inputs for selection, while \citet{hassid2024larger} show unit-test-based selection over many small-model samples can outperform one large-model sample. \citet{gao2025trae} bring test-time scaling to repository-level SWE issue resolution by combining ensemble reasoning with repo-aware exploration. SWE-Replay~\citep{ding2026swe} reduces the cost of naive scaling by branching from prior trajectories at carefully selected intermediate steps rather than sampling from scratch, and \citet{kim2026scaling} propose a representation-centric framework using compact trajectory summaries. These works largely treat context management and test-time scaling as separate axes; we instead study how an agent equipped with \textsc{CliffCompaction}{} responds to additional inference-time compute.

\paragraph{Continual learning for agents.}
Improving coding agents’ ability to continually learn from their own experience has emerged as an important research direction. Some work trains agents to improve their solutions through iterative refinement~\citep{baronio2026kevin, dai2026cuda}. Other approaches store prior experiences in external memory and retrieve relevant examples or strategies to guide the agent on new tasks~\citep{NEURIPS2023_1b44b878, wu2026from, qian-etal-2024-experiential, NEURIPS2024_0142921f, ouyang2026reasoningbank}. Evolutionary approaches maintain populations of candidate programs and use LLMs to propose and select mutations over successive iterations~\citep{novikov2025alphaevolvecodingagentscientific, li2025cocoevo}. However, many of these methods rely on task-specific training, memory structures, or optimization procedures, limiting their generalizability to new tasks.

%% file: contents/conclusion.tex


The benefits \textsc{CliffCompaction} brings depend on the scaffold and on the task. Across our experiments, cost savings grow with the complexity of the scaffold itself, and so does the budget the scaffold needs in order to work: the system prompt, tool definitions, and other fixed components consume part of the budget before the agent has taken a single action. Thus, the minimum threshold that an agent needs varies between scaffolds. Similarly, the benefit is meaningful only for medium-to-long-horizon tasks. 

Context management is a broad problem that has been approached in many different ways. Our comparison is limited to methods that act on the conversation history itself at inference time. We do not compare against approaches that train the model to manage its own context, or that maintain an external memory store the agent writes to and retrieves from. Such methods address the same problem from a different direction, and we did not have the resources to compare against them.

%% file: contents/efficiencyanalysis.tex
\subsection{Cost Savings with CliffCompaction}
\label{sec:cost}

Since cost is a primary consideration in real-world agent deployments, we assess the efficiency of \textsc{CliffCompaction} through cost. Tables~\ref{tab:swebench-verified} and \ref{tab:terminal-bench} report the average cost per instance.

Although API providers report token usage and cache statistics during agent execution, we observed substantial variation in cache-hit accounting across providers. To ensure a fair comparison, we recompute all costs using a perfect-caching model that derives cache hits directly from prompt lengths rather than provider-reported $\texttt{cached\_tokens}$ fields. Given a sequence of prompt lengths across API calls, $[P_1, P_2, \ldots, P_N]$, the first call is treated as a cold start with all $P_1$ tokens counted as non-cached. For each subsequent call $k > 1$, if $P_k \ge P_{k-1}$, we assume the previous prompt is fully cached, with $P_{k-1}$ cached tokens and $P_k - P_{k-1}$ non-cached tokens. In contrast, if $P_k < P_{k-1}$, we treat the reduction in prompt length as a compaction event that invalidates the KV cache, and all $P_k$ tokens are counted as non-cached. Completion tokens are not subject to caching and are counted per call at each provider's output price. This deterministic, model-agnostic procedure ensures that identical prompt sequences produce identical cache-hit rates regardless of provider, allowing us to isolate the cost effects of compaction from provider-specific caching behavior.


\paragraph{Overall Cost Savings.}
\textsc{CliffCompaction} consistently reduces cost across models, scaffolds, benchmarks, and context budgets. The magnitude of the savings depends on the model, with GLM generally benefiting more than Kimi. Savings also depend on the scaffold: on SWE-bench Verified, the gap between limited-context and unlimited-context settings is far larger for \texttt{OpenHands} than for \texttt{mini-swe-agent} (25.8\%--29.7\% versus 6.8\%--19.1\% at the same 32K budget). Benchmark-wise, the largest savings are observed on Terminal-Bench, where \textsc{CliffCompaction} reduces cost by 25.0\%--52.1\% for Kimi K2.6 and 35.8\%--64.6\% for GLM 5.1, substantially exceeding the reductions observed on SWE-bench Verified. 
We also compare against \texttt{Terminus-2} agent-based summarization on Kimi K2.6 (Table~\ref{tab:terminal-bench}). At a 32K budget, both approaches incur similar costs because compaction is triggered infrequently. Under tighter budgets, compaction kicks in at a substantially higher rate, and the additional summarization calls introduce non-trivial overhead: at 8K, \texttt{Terminus-2} summarization costs \$0.60 per instance against \$0.25 for \textsc{CliffCompaction}, while also resolving fewer tasks. We compare against further compaction baselines on long-horizon trajectories in Section~\ref{sec:compaction_comparison}.

\input{tables/token}

\paragraph{Cost Savings Breakdown.} A well-designed context-management strategy should not let re-prefilling costs overshadow the cache savings it unlocks. By decomposing the cost, we show that \textsc{CliffCompaction} strikes this balance. Table~\ref{tab:cost-attribution} shows that the savings achieved by \textsc{CliffCompaction} come entirely from reducing cache-read costs. Although cached tokens are substantially cheaper than uncached input or output tokens on a per-token basis, coding agents repeatedly reread their entire context throughout a trajectory, so cache reads dominate total cost, accounting for 78\% of the uncompacted bill. By limiting context growth, \textsc{CliffCompaction} directly targets this component, cutting cache-read cost by 80\%, from \$0.427 to \$0.087 per task. Compaction does unavoidably introduce increases in uncached input cost (from re-prefilling the compacted context) and output cost (from the additional turns following compaction), but these overheads remain minor: together they grow by \$0.06 per task, against \$0.34 in cache-read savings. Consequently, the cache-read reduction overwhelmingly outweighs these additional expenses, halving total cost.



\input{tables/real_cost}


Complementing the idealized perfect-cache model above, Table~\ref{tab:real-cost-cachehit} reports the actual observed costs together with cache hit rates. Overall, Kimi K2.6 attains higher and more stable cache hit rates (86--98\%) than GLM 5.1 (47--91\%), and the overall savings trends remain consistent with the perfect-cache model.

The cache hit rate explains how closely the real costs track the idealized estimates. For Kimi K2.6, whose cache hit rates are near-perfect, the real costs match the perfect-cache estimates to within ${\sim}1\%$. For GLM 5.1, the lower cache hit rate makes the real costs higher than the idealized estimates; consequently, the \emph{real} savings from \textsc{CliffCompaction} can exceed those predicted by the perfect-cache model. In both cases, cache hit rates decline steadily as the compaction threshold tightens (e.g., from 79\% to 47\% for GLM 5.1 on Terminal-Bench), reflecting the additional cache invalidation and re-prefilling induced by more frequent compaction.

The largest difference between the two models appears on SWE-bench Verified with \texttt{mini-swe-agent}. For GLM 5.1, cost reductions improve steadily as the threshold decreases, from 22\% at 32K to 58\% at 8K. Kimi K2.6, by contrast, is essentially unchanged at 32K, achieves a modest 8\% reduction at 16K, and is marginally more expensive at 8K---the additional re-prefilling at the tightest budget cancels the context savings for this already cost-efficient model. The same non-monotonic pattern appears on Terminal-Bench, where the 8K setting yields smaller savings (37\%) than 16K (53\%) for Kimi K2.6.

The magnitude of the savings also varies across scaffolds. On SWE-bench Verified, \texttt{OpenHands} benefits substantially more from \textsc{CliffCompaction} than \texttt{mini-swe-agent}: at a 32K threshold, both models save roughly 28--35\% on \texttt{OpenHands}, whereas savings on \texttt{mini-swe-agent} range from negligible to 22\%. The reductions on Terminal-Bench are the most pronounced, reaching as high as 72\% and remaining above 25\% across all configurations.

Model prices are provided in Table~\ref{tab:model-prices}.

\input{tables/price}

%% file: tables/token.tex
\begin{table}[!htbp]
  \centering
  \caption{Token and cost attribution by context component, before and after
compaction (Terminal-Bench 2.0, GLM 5.1). \emph{Other} combines uncached input with output.}
  \label{tab:cost-attribution}
  \footnotesize
  \setlength{\tabcolsep}{5pt}
  \begin{tabular}{l c cc c cc c}
    \toprule
    & & \multicolumn{3}{c}{No compaction} & \multicolumn{3}{c}{\textsc{CliffCompaction} @16K} \\
    \cmidrule(lr){3-5} \cmidrule(lr){6-8}
    Component & \% of Context & Cached & Other & Total & Cached & Other & Total \\
    \midrule
    Tool results  & 56.0\% & \$0.237 & \$0.029 & \$0.267 & \$0.053 & \$0.050 & $\$0.103_{\,(-61\%)}$ \\
    Tool calls    & 28.0\% & \$0.127 & \$0.062 & \$0.189 & \$0.012 & \$0.067 & $\$0.079_{\,(-58\%)}$ \\
    Thoughts      & 13.6\% & \$0.054 & \$0.030 & \$0.084 & \$0.013 & \$0.062 & $\$0.075_{\,(-11\%)}$ \\
    System \& task&  2.4\% & \$0.008 & \$0.001 & \$0.009 & \$0.010 & \$0.001 & $\$0.011_{\,(+22\%)}$ \\
    \midrule
    Total         &        & \$0.427 & \$0.121 & \$0.548 & \$0.087 & \$0.181 & $\$0.268_{\,(-51\%)}$ \\
    \bottomrule
  \end{tabular}
\end{table}

%% file: tables/real_cost.tex
  \begin{table}[!htbp]
      \centering
      \caption{Real (provider-metered) cost and provider cache-hit rate across scaffolds and context budgets. Subscripts on \emph{Real Cost} give the cost change relative to the Full-context baseline (negative = cheaper, positive = more expensive).}
      \label{tab:real-cost-cachehit}
      \footnotesize
      \setlength{\tabcolsep}{4pt}
      \resizebox{\textwidth}{!}{%
      \begin{tabular}{ll cc cc cc cc}
        \toprule
        & & \multicolumn{2}{c}{Full-context} & \multicolumn{2}{c}{32k}
        & \multicolumn{2}{c}{16k} & \multicolumn{2}{c}{8k} \\
        \cmidrule(lr){3-4} \cmidrule(lr){5-6} \cmidrule(lr){7-8} \cmidrule(lr){9-10}
        Scaffold & Model & Real Cost & Cache Hit & Real Cost & Cache Hit
                 & Real Cost & Cache Hit & Real Cost & Cache Hit \\
        \midrule
        \addlinespace[2pt]
        \multicolumn{10}{@{}c}{\texttt{SWE-bench Verified}} \\
        \midrule
        \multirow{2}{*}{\texttt{mini-swe-agent}}
          & Kimi K2.6 & \$0.19 & 96\% & \$0.19$_{+0\%}$  & 95\% & \$0.17$_{-8\%}$  & 92\% & \$0.19$_{+1\%}$  & 86\% \\
          & GLM 5.1   & \$0.38 & 82\% & \$0.29$_{-22\%}$ & 81\% & \$0.19$_{-50\%}$ & 84\% & \$0.16$_{-58\%}$ & 76\% \\
        \midrule
        \multirow{2}{*}{\texttt{OpenHands}}
          & Kimi K2.6 & \$0.57 & 98\% & \$0.41$_{-28\%}$ & 96\% & ---    & ---  & ---    & ---  \\
          & GLM 5.1   & \$1.27 & 91\% & \$0.83$_{-35\%}$ & 90\% & ---    & ---  & ---    & ---  \\
        \midrule
        \addlinespace[2pt]
        \multicolumn{10}{@{}c}{\texttt{Terminal-Bench 2.0}} \\
        \midrule
        \multirow{2}{*}{\texttt{terminus-2}}
          & Kimi K2.6 & \$0.40 & 97\% & \$0.30$_{-26\%}$ & 94\% & \$0.19$_{-53\%}$ & 87\% & \$0.25$_{-37\%}$ & 79\% \\
          & GLM 5.1   & \$0.89 & 79\% & \$0.53$_{-40\%}$ & 68\% & \$0.40$_{-55\%}$ & 58\% & \$0.25$_{-72\%}$ & 47\% \\
        \bottomrule
      \end{tabular}%
      }
  \end{table}

%% file: tables/price.tex
\begin{table}[!htbp]
  \centering
  \caption{Token prices used for cost calculation. Prices are reported in USD per one million tokens.}
  \label{tab:model-prices}
  \footnotesize
  \setlength{\tabcolsep}{5pt}
  \begin{tabular}{lccc}
    \toprule
    Model & Input & Cache read & Output \\
    \midrule
    GLM 4.7 Flash & \$0.00 & \$0.00 & \$0.00 \\
    GLM 5         & \$1.00 & \$0.20 & \$3.20 \\
    GLM 5 Turbo   & \$1.20 & \$0.24 & \$4.00 \\
    GLM 5.1       & \$1.40 & \$0.26 & \$4.40 \\
    GLM 5.2       & \$1.40 & \$0.26 & \$4.40 \\
    GLM 5.3 Flash & \$0.15 & \$0.03 & \$0.50 \\
    Kimi K2.5     & \$0.60 & \$0.10 & \$3.00 \\
    Kimi K2.6     & \$0.95 & \$0.16 & \$4.00 \\
    Kimi K2.7     & \$0.95 & \$0.19 & \$4.00 \\
    \bottomrule
  \end{tabular}
  \vspace{2pt}
\end{table}


%% file: contents/extended_efficiency.tex
\subsection{Measured Runtime}
\label{appendix:runtime}
We also report runtime as an additional axis for evaluating the efficiency gains of \textsc{CliffCompaction}. We use the per-instance wall-clock that \texttt{OpenHands} logs directly on SWE-bench Verified, the setting that also sustains the highest, most stable cache-hit rates (above 90\%). As shown in Table~\ref{tab:duration}, with the context window capped at 32K, \textsc{CliffCompaction} reduces per-instance wall-clock time by 23\% and 34\% for Kimi K2.6 and GLM 5.1, respectively---closely tracking the corresponding cost reductions of 28\% and 35\%. This indicates that the benefits of \textsc{CliffCompaction} extend beyond monetary cost to end-to-end execution time.

Table~\ref{tab:glm-flash-time} shows the run time and speedup of \textsc{CliffCompaction} on \texttt{mini-swe-agent} with GLM-4.7-Flash served by vLLM (RTX PRO 6000 Blackwell). The same pattern holds: inference time falls monotonically as the threshold tightens, giving speedups of $1.60\times$, $2.10\times$ and $2.62\times$ at 32K, 16K and 8K. This highlights that the benefits of \textsc{CliffCompaction} extend to self-hosted serving setups.

\input{tables/duration}

\subsection{\textsc{CliffCompaction} vs.\ Other Compaction Methods}
\label{appendix:swe-compaction}

\begin{figure}[ht]
  \centering
  \includegraphics[width=\linewidth]{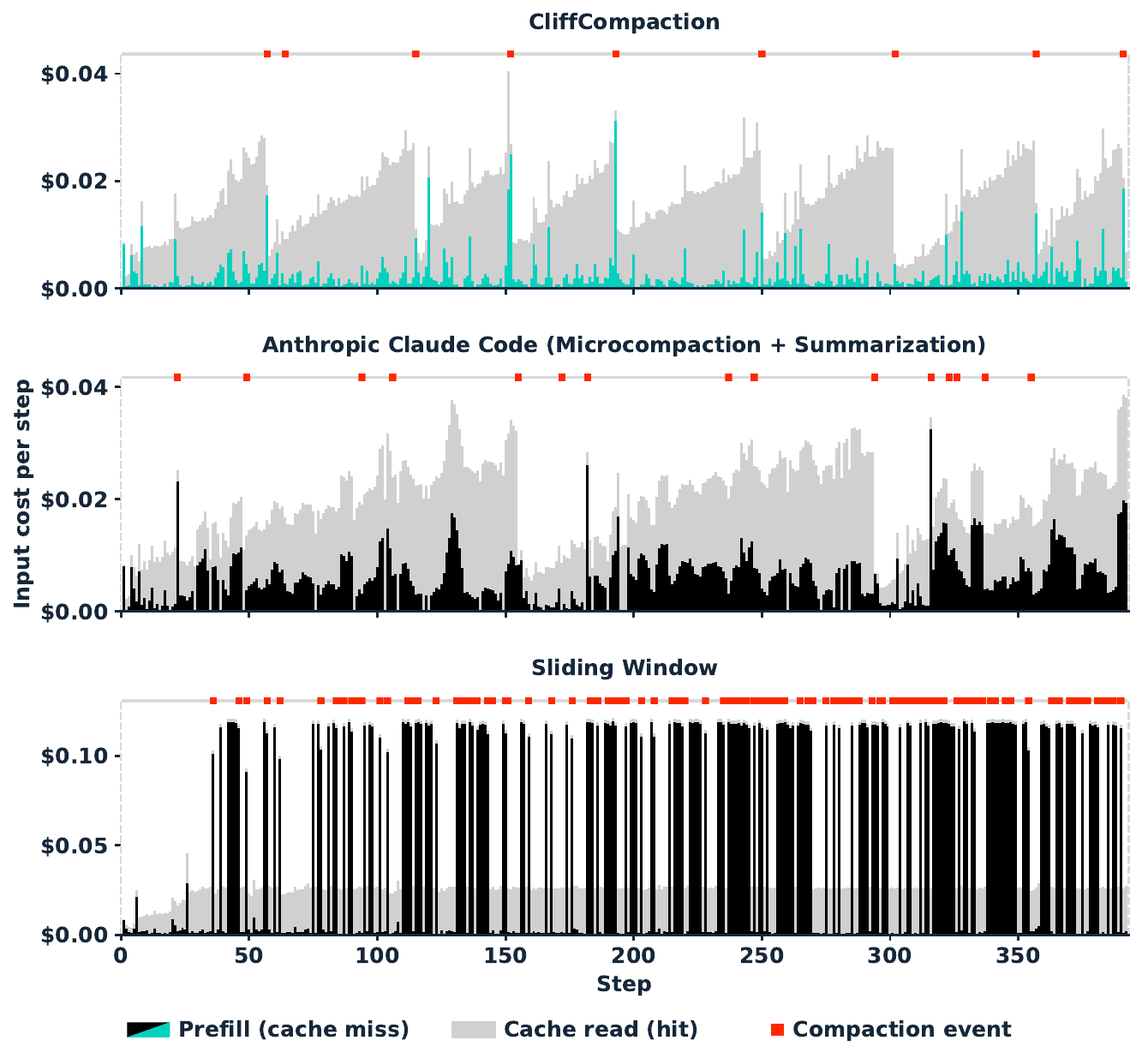}
  \caption{Per-step input cost on a KernelBench task, decomposed into uncached prefill and cache reads.}
  \label{fig:prefill-kernelbench}
\end{figure}

\begin{figure}[t]
  \centering
  \includegraphics[width=0.8\linewidth]{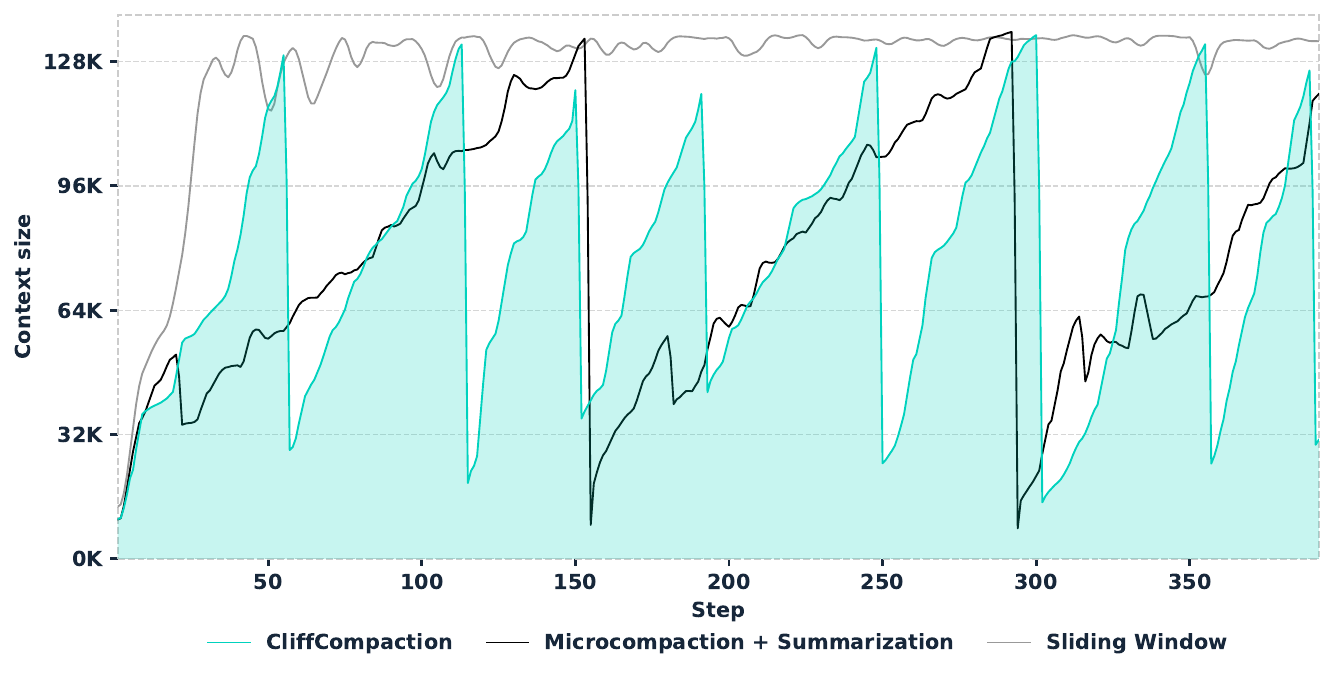}\\[4pt]
  \caption{Context size over steps on a KernelBench task.}
  \label{fig:context-kernelbench}
\end{figure}

Figures~\ref{fig:prefill-kernelbench} and~\ref{fig:context-kernelbench} trace a single representative KernelBench task step by step. The context trace (Figure~\ref{fig:context-kernelbench}) is the real-data counterpart of Figure~\ref{fig:cliff}. \textsc{CliffCompaction} grows its context append-only and drops it in a clean cliff at each compaction. Sliding window holds the context at its ceiling. Microcompaction~+~summarization never lets the context settle, continually masking and rewriting old observations, producing the sawtooth teeth that ripple through the entire trajectory.

These patterns produce the per-step costs in Figure~\ref{fig:prefill-kernelbench}, which follow exactly the cost ordering anticipated in Figure~\ref{fig:cliff}: cheap for \textsc{CliffCompaction}, intermediate for microcompaction~+~ summarization, and expensive for sliding window. Because \textsc{CliffCompaction} leaves earlier tokens untouched between compactions, its context is served almost entirely from cache (gray), and it pays an uncached re-prefill (colored) only at its rare resets. Microcompaction's constant prefix edits invalidate the KV-cache again and again, making every tooth a fresh partial re-prefill, so the cache never has a chance to amortize and the per-step cost cannot fall far. Sliding window is the worst, shifting the prefix at nearly every step and re-prefilling close to the full context each time. \textsc{CliffCompaction} thus remains the most efficient compaction method.

%% file: tables/duration.tex
\begin{table}[!htbp]
    \centering
    \caption{Per-instance wall-clock duration on \texttt{OpenHands} (SWE-bench Verified), with the corresponding cost reduction.}
    \label{tab:duration}
    \footnotesize
    \setlength{\tabcolsep}{6pt}
    \begin{tabular}{l cc cc}
      \toprule
      & \multicolumn{2}{c}{Duration} & \multicolumn{2}{c}{Reduction at 32k} \\
      \cmidrule(lr){2-3} \cmidrule(lr){4-5}
      Model & Full-context & 32k & Duration & Cost \\
      \midrule
      Kimi K2.6 & 849\,s & 656\,s & \cellcolor{cliffteal!44}{$-23\%$} & \cellcolor{cliffteal!44}{$-28\%$} \\
      GLM 5.1   & 900\,s & 592\,s & \cellcolor{cliffteal!44}{$-34\%$} & \cellcolor{cliffteal!44}{$-35\%$} \\
      \bottomrule
    \end{tabular}
\end{table}

\begin{table}[!htbp]
  \centering
  \caption{Measured inference time of \textsc{CliffCompaction} at different context thresholds (GLM-4.7-Flash, vLLM on RTX PRO 6000 Blackwell, SWE-bench Verified). Time is the total latency spent in LLM calls. Speedup is relative to the full-context run.}
  \label{tab:glm-flash-time}
  \footnotesize
  \begin{tabular}{lcc}
    \toprule
    Threshold & Total time (h) & Speedup \\
    \midrule
    Full context & 234.2 & --- \\
    32K & 146.0 & \cellcolor{cliffteal!44}$1.60\times$ \\
    16K & 111.6 & \cellcolor{cliffteal!44}$2.10\times$ \\
    \phantom{0}8K & \phantom{0}88.9 & \cellcolor{cliffteal!44}$2.62\times$ \\
    \bottomrule
  \end{tabular}
\end{table}

%% file: contents/extended_cliff.tex



\textsc{CliffCompaction} keeps literal fragments of the original context, truncating them rather than rewriting or paraphrasing their content. For each action, it keeps only a lightweight signature: the first 150 characters of the action string. In \texttt{mini-swe-agent} and \texttt{Terminus-2}---whose actions are primarily bash commands and terminal keystrokes, respectively---this signature is the leading 150 characters of each command (the bash command parsed from the model's \texttt{bash} code block in \texttt{mini-swe-agent-v1}, and the \texttt{keystrokes} parsed from the model's response in \texttt{Terminus-2}). This simple truncation captures most of the useful information, since the command itself typically appears at the start of the string, while the remainder often consists of lengthy inline file contents. For \texttt{OpenHands}, the tool-call signatures are listed in Table~\ref{tab:openhands-signatures}.

\input{tables/oh_tool_signature}

%% file: tables/oh_tool_signature.tex
\begin{table}[t]
\centering
\caption{Tool-call signatures used by \textsc{CliffCompaction} in \texttt{OpenHands}
scaffold.}
\small
\begin{tabular}{@{}llp{0.40\linewidth}@{}}
\toprule
\textbf{Tool} & \textbf{Sub-command} & \textbf{Retained signature} \\
\midrule
\texttt{terminal}    & ---          & \texttt{[terminal] \{command\}} \\
\addlinespace
\texttt{file\_editor} & \texttt{view}        & \texttt{[file\_editor view] \{path\} range=\{range\}} \\
\texttt{file\_editor} & \texttt{create}      & \texttt{[file\_editor create] \{path\} (\{N\} chars)} \\
\texttt{file\_editor} & \texttt{str\_replace} & \texttt{[file\_editor str\_replace] \{path\} (old: \{old\_str\}...)} \\
\texttt{file\_editor} & \texttt{insert}      & \texttt{[file\_editor insert] \{path\}:\{line\}} \\
\texttt{file\_editor} & \texttt{undo\_edit}  & \texttt{[file\_editor undo\_edit] \{path\}} \\
\addlinespace
\texttt{think}       & ---          & \texttt{[think] \{thought\}} \\
\texttt{finish}      & ---          & \texttt{[finish]} \\
\texttt{task\_tracker} & ---        & \text{(dropped)} \\
\midrule
\multicolumn{3}{@{}l@{}}{\footnotesize Caps: \texttt{command} $\le 120$, \texttt{old\_str} $\le 60$ chars.} \\
\bottomrule
\end{tabular}
\label{tab:openhands-signatures}
\end{table}

%% file: contents/extended_analysis.tex
\label{appendix:extended_analysis}

\subsection{\textsc{CliffCompaction} has high precision}
\label{appendix:precision}

\begin{figure}[!htbp]
  \centering
  \includegraphics[width=\linewidth]{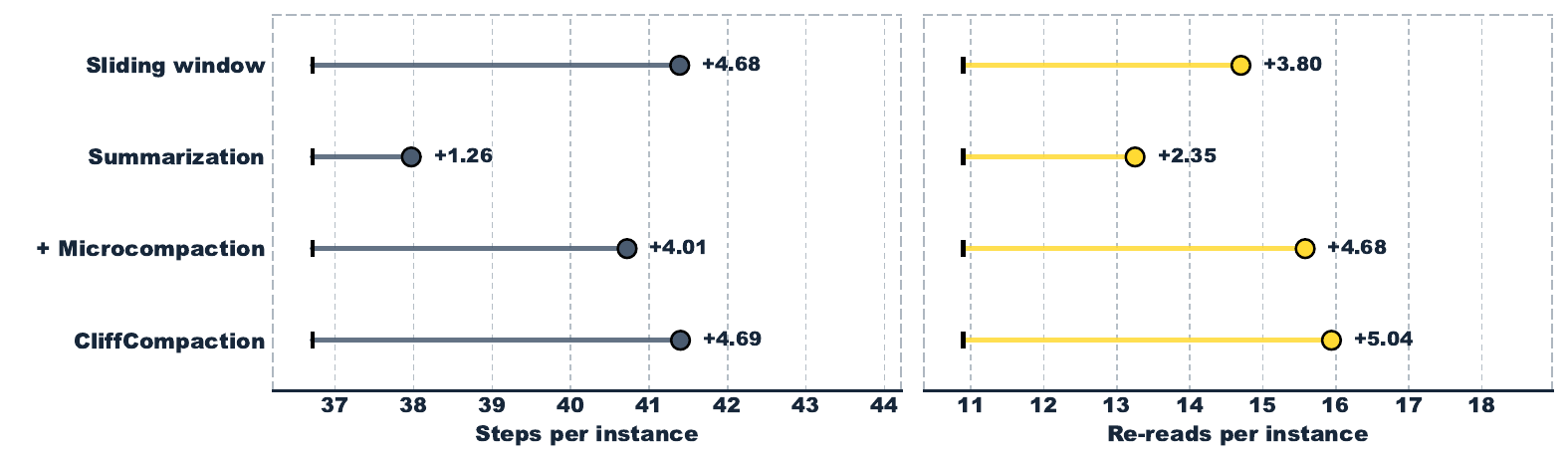}
  \caption{$\Delta$ of mean actions per instance and steps spent on re-reads (SWE-bench Verified, Kimi K2.7, $16K$) relative to running without compaction (full-context: $36.71$ steps, $10.90$ re-reads per instance).}
  \label{fig:extra-steps}
\end{figure}

We justify the precision hypothesis behind \textsc{CliffCompaction} through the agent's re-reading behaviour. Figure~\ref{fig:extra-steps} shows that every compaction strategy increases the number of steps per trajectory relative to the full-context runs, with sliding window, summarization + microcompaction and \textsc{CliffCompaction} each adding more than four steps, while summarization rises by only $+1.26$.

Almost all of these additional steps are spent re-reading files the agent had already opened. Editing and testing, by contrast, decrease under every strategy. \textsc{CliffCompaction}, sliding window and microcompaction discard tool outputs outright, leaving a visible gap in the context, so the agent re-fetches the content when it needs it again ($+5.04$, $+3.80$ and $+4.68$ re-reads respectively). Summarization adds the fewest re-reads ($+2.35$) and is the only strategy that also opens fewer new files ($-0.68$ per instance). It replaces the discarded content with a plausible natural-language paraphrase, which appears to satisfy the agent and discourages it from returning to the ground-truth source.

This behaviour corroborates the low-recall, high-precision trade-off we posit for \textsc{CliffCompaction}: it retains less than summarization, but what it retains is verbatim, and the gaps it leaves are unambiguous, so the agent restores what it needs from the source.

\subsection{Test-time scaling on SWE-bench Verified}

\begin{table}[!htbp]
  \centering
  \caption{Test-time scaling on SWE-bench Verified, ordered by accuracy. All rows use \texttt{mini-swe-agent} except CodeMonkeys, which uses its own.}
  \label{tab:swebench-scaling}
  \footnotesize
  \setlength{\tabcolsep}{5pt}
  \resizebox{\textwidth}{!}{%
  \begin{tabular}{l l c r c c c}
    \toprule
    Model & Context & $k$ & Cost (USD) & Oracle & Practical & Gain\,@\,Cost\textsuperscript{*} \\
    \midrule
    \rowcolor{cliffteal!44}
    Kimi K2.6 \textsc{+ SGV} & 256K & 5 & \$473.80 & 83.0 & 79.4 & +4.8\,@\,5.0$\times$ \\
    \rowcolor{citepurple!12}
    LLM-as-a-Verifier\textsuperscript{\S} & mixed & 3 & $\sim$\$590.00 & 84.4 & 78.2 & +2.1\,@\,3.2$\times$ \\
    \rowcolor{cliffteal!44}
    Kimi K2.6 \textsc{+ SGV} & 256K & 3 & \$284.28 & 80.4 & 78.0 & +3.4\,@\,3.0$\times$ \\
    Kimi K2.6 \textsc{+ Cliff + SGV} & 16K & 5 & \$440.10 & 80.8 & 77.0 & +2.4\,@\,4.64$\times$ \\
    \rowcolor{citepurple!12}
    Opus 4.5 (high reasoning)  & 200K & 1 & \$375.00 & --- & 76.8 & --- \\
    GLM 5.1 \textsc{+ SGV} & 200K & 5 & \$604.65 & 80.6 & 76.6 & +5.8\,@\,5.0$\times$ \\
    \rowcolor{cliffteal!44}
    Kimi K2.6 \textsc{+ Cliff + SGV} & 16K & 3 & \$264.06 & 79.4 & 76.4 & +1.8\,@\,2.79$\times$ \\
    Kimi K2.6 \textsc{+ Cliff + SGV} & 32K & 5 & \$459.60 & 82.8 & 76.2 & +1.6\,@\,4.85$\times$ \\
    Kimi K2.6 \textsc{+ Cliff + SGV} & 32K & 3 & \$275.76 & 79.8 & 76.0 & +1.4\,@\,2.91$\times$ \\
    GLM 5.1 \textsc{+ SGV} & 200K & 3 & \$362.79 & 78.0 & 75.8 & +5.0\,@\,3.0$\times$ \\
    \rowcolor{citepurple!12}
    Gemini 3 Flash (high reasoning) & 1M & 1 & \$180.00 & --- & 75.8 & --- \\
    \rowcolor{citepurple!12}
    MiniMax M2.5 (high reasoning) & 205K & 1 & \$35.00 & --- & 75.8 & --- \\
    \rowcolor{citepurple!12}
    Opus 4.6                   & 1M & 1 & \$275.00 & --- & 75.6 & --- \\
    GLM 5.1 \textsc{+ Cliff + SGV} & 32K & 5 & \$519.85 & 79.2 & 74.8 & +4.0\,@\,4.30$\times$ \\
    Kimi K2.6                  & 256K & 1 & \$94.76  & --- & 74.6 & --- \\
    \rowcolor{cliffteal!44}
    GLM 5.1 \textsc{+ Cliff + SGV} & 16K & 3 & \$229.89 & 78.6 & 74.0 & +3.2\,@\,1.90$\times$ \\
    GLM 5.1 \textsc{+ Cliff + SGV} & 32K & 3 & \$311.91 & 76.2 & 74.0 & +3.2\,@\,2.58$\times$ \\
    GLM 5.1 \textsc{+ Cliff + SGV} & 16K & 5 & \$383.15 & 81.4 & 73.8 & +3.0\,@\,3.17$\times$ \\
    GLM 5.1                    & 200K & 1 & \$120.93 & --- & 70.8 & --- \\
    \rowcolor{citepurple!12}
    CodeMonkeys (Sonnet 3.5 + Qwen 2.5)\textsuperscript{\P} & 200K & 10 & \$2291.90 & 69.8 & 57.4 & --- \\
    \rowcolor{citepurple!12}
    Sonnet 3.7      & 200K & 1 & \$175.00 & --- & 52.8 & --- \\
    \bottomrule
  \end{tabular}%
  }

  \vspace{3pt}
  \begin{minipage}{\textwidth}
    \footnotesize
    \setlength{\parindent}{0pt}
    \textsuperscript{*}\,Accuracy gain and cost multiplier relative to a single uncompacted rollout ($k{=}1$).\par
    \textsuperscript{\S}\,From \citet{kwok2026llmasaverifiergeneralpurposeverificationframework}; Gemini-2.5-Flash as the verifier. Three candidates, each generated by one of Claude Opus 4.5, Gemini 3 Flash, and MiniMax M2.5. Cost: generation \$375$+$\$180$+$\$35 $\approx$ \$590.\par
    \textsuperscript{\P}\,From \citet{ehrlich2025codemonkeysscalingtesttimecompute}.\par
  \end{minipage}
\end{table}

On SWE-bench Verified (Table~\ref{tab:swebench-scaling}), Kimi K2.6 with \textsc{SGV} reaches the highest Practical accuracy in the table (79.4\%), topping every proprietary single-model baseline, including the strongest, Opus 4.5 (76.8\%). It also exceeds the LLM-as-a-Verifier (78.2\%), and comes within $0.2$ points of it at $k{=}3$ at roughly half the cost (\$284.28 vs.\ \$625), without the repeated LLM-verifier calls their method requires.

With \textsc{CliffCompaction}, capping context at a lower threshold widens this cost advantage. Three compacted Kimi rollouts at 16K reach 76.4\% for \$264.06, within $0.4$ points of Opus 4.5 at $70\%$ of its cost, and five reach 77.0\%, surpassing it. On cost, GLM 5.1 at 16K is the cheapest scaled configuration overall (74.0\% at \$229.89). As a result, the cheapest Pareto-optimal scaled configurations are compacted ones.

\subsection{Speedup by Kernels Generated on KernelBench}
\label{appendix:kernels_steps}

\begin{figure}[t]
  \centering
  \begin{subfigure}[t]{0.49\textwidth}
    \centering
    \includegraphics[width=\linewidth]{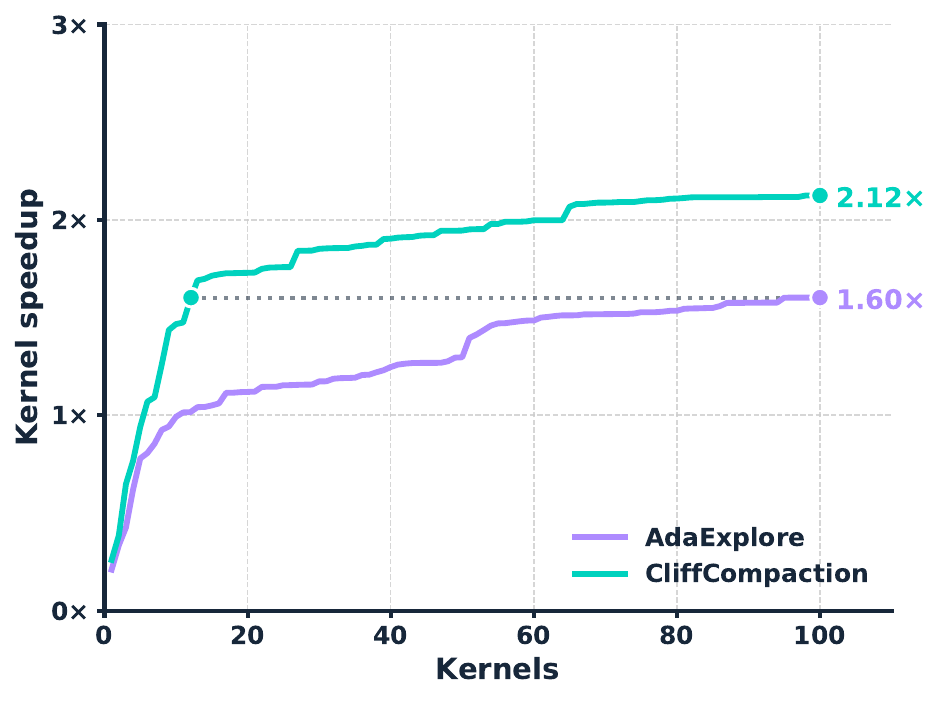}
    \caption{\textsc{CliffCompaction} vs.\ AdaExplore.}
    \label{fig:kb-kernel-ada}
  \end{subfigure}
  \hfill
  \begin{subfigure}[t]{0.49\textwidth}
    \centering
    \includegraphics[width=\linewidth]{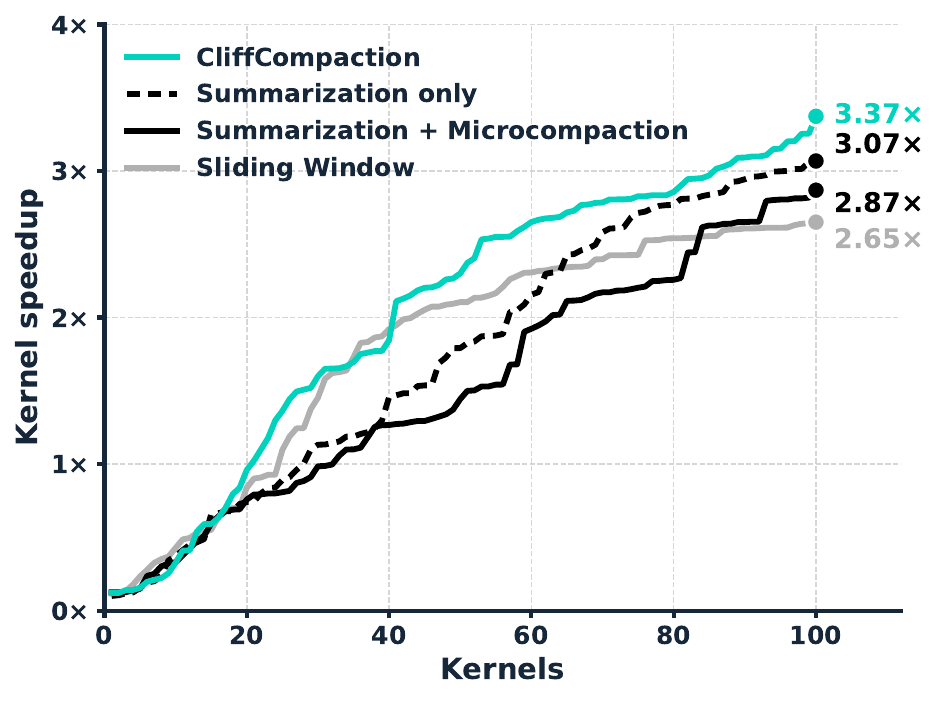}
    \caption{\textsc{CliffCompaction} vs.\ compaction alternatives.}
    \label{fig:kb-kernel-compaction}
  \end{subfigure}
  \caption{Best-so-far kernel speedup on KernelBench Level~3, indexed by kernels generated, AdaExplore's budget unit.}
  \label{fig:kb-kernel-curves}
\end{figure}

Search-based methods such as AdaExplore budget by candidate kernels: each of their steps is one kernel proposal. We therefore also report speedup indexed by kernels generated (Figure~\ref{fig:kb-kernel-curves}), measuring which method
finds the faster kernel given the same number of candidates to compile and benchmark. Because our setup is agentic, we count a kernel as an agent turn that writes compiled kernel code.

Figure~\ref{fig:kb-kernel-ada} shows that once exploration steps are excluded, \textsc{CliffCompaction} leads AdaExplore at every budget, with a clear margin by the tenth kernel ($1.47\times$ against $0.99\times$). It reaches $1.60\times$ after 13 kernels, a level AdaExplore takes 100 kernels to reach. This again shows that a simple agentic setup with autocompaction compares favorably with a pipeline built specifically for kernel optimization.

Figure~\ref{fig:kb-kernel-compaction} shows the same advantage over the other compaction baselines. Within the same step budget (Table~\ref{tab:compaction-cost}), \textsc{CliffCompaction} generates the fewest candidates, so normalizing by proposals widens the gap. Summarization spends fewer steps re-reading ($91.8$ reads per problem vs.\ $102.7$) and more steps writing kernels, so its proposals are individually less effective. \textsc{CliffCompaction}, in contrast, re-reads from the workspace and converts each proposal into more speedup, the same pattern appears on SWE-bench (Appendix~\ref{appendix:precision}).

%% file: contents/extended_setup.tex
\subsection{Additional Test-Time Scaling Details}
\label{appendix:tts}

In this subsection, we describe the selector used to obtain the \textit{Practical} results in Table~\ref{tab:terminalbench-scaling} and Table~\ref{tab:swebench-scaling}.

Each candidate is represented by a 30-dimensional feature vector (Table~\ref{tab:tts-features} and Table~\ref{tab:tts-features-swebench}). We train a LightGBM binary classifier: given the feature vector of a candidate trajectory, it predicts the probability that the candidate resolves the task, and a separate classifier is trained for each rollout budget $k$. We use the following hyperparameters on both benchmarks: $\{$800 boosting rounds, learning rate $0.03$, 63 leaves, minimum 15 samples per leaf, $\ell_2$ regularization $1.0\}$. We train the selector on rollouts from a different model than the one it is applied to: the selector used on Kimi K2.6 rollouts is trained only on GLM 5.1 rollouts, and vice versa. Training on rollouts from the same model gives no consistent advantage (Table~\ref{tab:tts-same-vs-cross}).

We also examine the task-generalization ability of the selector. We train it on a fraction of the tasks and evaluate on all 89 Terminal-Bench 2.0 tasks (Table~\ref{tab:tts-data-efficiency}). Across training fractions from 25\% to 100\%, the resolution rate remains essentially unchanged for every configuration. This shows that the selector does not overfit to the specific training tasks, it generalizes across tasks and performs comparably even when trained on only a quarter of them. In the main results (Table~\ref{tab:terminalbench-scaling}) we use the selector trained on 100\% of the tasks.

\begin{table}[!htbp]
  \centering
  \caption{The feature set used by \textsc{SGV} on Terminal-Bench 2.0. $\Delta\mu$ denotes the feature minus its within-group mean; $/\max$ denotes the feature divided by its within-group maximum. These instance-normalized variants make a candidate's value relative to the others of the same task.}
  \label{tab:tts-features}
  \footnotesize
  \setlength{\tabcolsep}{6pt}
  \begin{tabular}{l l}
    \toprule
    Feature & Description \\
    \midrule
    \multicolumn{2}{@{}l}{\textit{Execution behavior}} \\
    \midrule
    \texttt{marked\_complete}            & Agent explicitly signaled task completion \\
    \texttt{n\_episodes}                 & Number of agent steps \\
    \texttt{n\_episodes} $\Delta\mu$     & \quad relative to group mean \\
    \texttt{n\_tool\_calls}              & Number of tool calls \\
    \texttt{n\_tool\_calls} $\Delta\mu$  & \quad relative to group mean \\
    \texttt{n\_tool\_calls} $/\max$      & \quad relative to group max \\
    \midrule
    \multicolumn{2}{@{}l}{\textit{Produced solution}} \\
    \midrule
    \texttt{n\_files}                    & Number of files written \\
    \texttt{n\_lines}                    & Non-trivial lines of code \\
    \texttt{n\_raw\_lines}               & Raw lines (incl.\ blanks/short) \\
    \texttt{n\_paths\_root}              & Files written under an absolute path \\
    \texttt{n\_unique\_idents}           & Distinct identifiers \\
    \texttt{is\_shell}                   & Solution is a shell script \\
    \texttt{avg\_line\_len}              & Mean line length \\
    \texttt{avg\_line\_len} $\Delta\mu$  & \quad relative to group mean \\
    \texttt{max\_line\_len}              & Maximum line length \\
    \texttt{std\_line\_len}              & Std.\ dev.\ of line length \\
    \texttt{max\_indent} $\Delta\mu$     & Max indentation depth, rel.\ to group mean \\
    \texttt{n\_indent\_levels} $\Delta\mu$ & Distinct indent levels, rel.\ to group mean \\
    \texttt{alnum\_ratio}                & Alphanumeric character ratio \\
    \texttt{alnum\_ratio} $\Delta\mu$    & \quad relative to group mean \\
    \texttt{punct\_density}              & Punctuation character density \\
    \texttt{kw\_if}                      & Count of \texttt{if}/\texttt{elif} \\
    \texttt{kw\_if} $\Delta\mu$          & \quad relative to group mean \\
    \texttt{kw\_while}                   & Count of \texttt{while} \\
    \texttt{kw\_print} $/\max$           & Count of \texttt{print}/\texttt{echo}, rel.\ to group max \\
    \texttt{kw\_shebang}                 & Count of shebang (\texttt{\#!}) lines \\
    \midrule
    \multicolumn{2}{@{}l}{\textit{Within-group agreement}} \\
    \midrule
    \texttt{overlap\_min}                & Min line overlap with other candidates \\
    \texttt{overlap\_sum} $\Delta\mu$    & Mean line overlap, rel.\ to group mean \\
    \texttt{symbol\_overlap\_sum}        & Shared identifiers with other candidates \\
    \texttt{symbol\_jaccard\_sum} $/\max$ & Symbol Jaccard agreement, relative to group max \\
    \bottomrule
  \end{tabular}
\end{table}

\begin{table}[!htbp]
  \centering
  \caption{The feature set used by \textsc{SGV} on SWE-bench Verified, for the selector applied to Kimi K2.6 rollouts (trained on GLM 5.1). $\Delta\mu$ denotes the feature minus its within-group mean; $/\max$ denotes the feature divided by its within-group maximum.}
  \label{tab:tts-features-swebench}
  \footnotesize
  \setlength{\tabcolsep}{6pt}
  \begin{tabular}{l l}
    \toprule
    Feature & Description \\
    \midrule
    \multicolumn{2}{@{}l}{\textit{Produced patch}} \\
    \midrule
    \texttt{n\_lines}                    & Non-trivial changed lines \\
    \texttt{n\_add}                      & Added lines \\
    \texttt{n\_del}                      & Deleted lines \\
    \texttt{n\_del} $\Delta\mu$          & \quad relative to group mean \\
    \texttt{n\_del} $/\max$              & \quad relative to group max \\
    \texttt{add\_del\_ratio}             & Ratio of added to deleted lines \\
    \texttt{n\_files}                    & Files modified \\
    \texttt{n\_hunks}                    & Diff hunks \\
    \texttt{patch\_func\_count}          & Distinct function names in \texttt{def} statements appearing in the patch \\
    \texttt{patch\_func\_count} $/\max$  & \quad relative to group max \\
    \texttt{n\_unique\_idents}           & Distinct identifiers \\
    \texttt{n\_total\_tokens}            & Total identifier tokens \\
    \texttt{ident\_repeat\_ratio}        & Identifier tokens per distinct identifier \\
    \texttt{avg\_line\_len}              & Mean line length \\
    \texttt{avg\_line\_len} $/\max$      & \quad relative to group max \\
    \texttt{max\_line\_len}              & Maximum line length \\
    \texttt{sig\_if}                     & Lines matching an \texttt{if} signature \\
    \texttt{sig\_return}                 & Lines matching a \texttt{return} signature \\
    \texttt{kw\_if}                      & Count of \texttt{if}/\texttt{elif} \\
    \texttt{kw\_import}                  & Count of \texttt{import} \\
    \midrule
    \multicolumn{2}{@{}l}{\textit{Repository context}} \\
    \midrule
    \texttt{repo\_code}                  & Repository identity (categorical) \\
    \midrule
    \multicolumn{2}{@{}l}{\textit{Within-group agreement}} \\
    \midrule
    \texttt{group\_consensus\_level}     & Mean pairwise line Jaccard across the task's candidates \\
    \texttt{group\_diversity}            & $1-$\texttt{group\_consensus\_level} \\
    \texttt{group\_has\_exact\_duplicates} & An identical patch occurs at least twice in the group \\
    \texttt{file\_agreement\_sum}        & Other candidates modifying the same files \\
    \texttt{symbol\_overlap\_sum}        & Changed identifiers shared with other candidates \\
    \texttt{add\_overlap\_sum}           & Added lines shared with other candidates \\
    \texttt{del\_overlap\_sum}           & Deleted lines shared with other candidates \\
    \texttt{hunk\_jaccard\_sum}          & Jaccard agreement of diff hunks with other candidates \\
    \texttt{max\_symbol\_jaccard}        & Symbol Jaccard with the most similar other candidate \\
    \bottomrule
  \end{tabular}
\end{table}

\begin{table}[!htbp]
  \centering
  \caption{Same-model versus cross-model training of \textsc{SGV} on Terminal-Bench (Kimi K2.6, $k{=}3$). We used $5$-fold instance-level cross-validation. $\Delta$ is same-model minus cross-model; neither shows a consistent advantage.}
  \label{tab:tts-same-vs-cross}
  \footnotesize
  \setlength{\tabcolsep}{6pt}
  \begin{tabular}{l c c c c c}
    \toprule
    Context & pass@1 & Oracle & Same-model & Cross-model & $\Delta$ \\
    \midrule
    256K & 59.2 & 70.8 & 61.8 & 60.7 & $+1.1$ \\
    16K  & 61.4 & 74.2 & 63.7 & 64.4 & $-0.7$ \\
    \bottomrule
  \end{tabular}
\end{table}

\input{tables/tts_data_efficiency}

\subsection{Additional KernelBench Details}
\label{appendix:kernelbench}

\paragraph{Agent setup.}
For the CUDA setting, we follow the KernelBench setup of \citet{dai2026cuda}, using the same skill and working environment. Each task is a self-contained CUDA C++ extension project in which the agent writes raw CUDA kernels and is barred from falling back on PyTorch compute operators, so reported speedups reflect genuinely hand-written kernels; we refer to \citet{dai2026cuda} for the full skill and agent loop. The agent is instructed to optimize the reference model for maximum speedup over the PyTorch Eager and \texttt{torch.compile} baselines and to keep proposing new optimizations. We differ from \citet{dai2026cuda} only in: (i)~the backbone model (Kimi~K2.7 and Kimi~K2.6 vs.\ their RL-trained model); (ii)~the hardware (L40S and RTX~Pro~6000 Blackwell vs.\ their H20); (iii)~the stopping condition---we remove the agent's ability to self-terminate and run to a fixed budget of steps (or, without compaction, until the 256K context window is exceeded); and (iv)~context management via \textsc{CliffCompaction}.

For the Triton setting, we run GPT-5-mini in the same environment but target Triton rather than CUDA and do not block PyTorch operators, matching
\citet{du2026adaexplore}.

\paragraph{Evaluation setup.}
To score the speedups achieved by the agent, we follow \citet{du2026adaexplore}. At each step, we evaluate the kernel the agent produces: we compile it, check that its output matches that of the reference model on 5 random inputs within a tolerance of $\texttt{atol}=\texttt{rtol}=10^{-2}$ for our CUDA configurations and $\texttt{atol}=\texttt{rtol}=5\times10^{-2}$ for our Triton configuration, and if it is correct, measure its runtime against the PyTorch Eager baseline (averaged over 10 timed runs after 10 warmup runs). For each problem we record the best speedup the agent achieves across all steps, clamped to a maximum of $10\times$. If the agent fails to produce any correct kernel within the step budget, that problem is penalized with a speedup of $0.1$. The reported \textit{Speedup} is the geometric mean of these per-problem speedups over all 50 tasks.

%% file: tables/tts_data_efficiency.tex
\begin{table}[!htbp]
  \centering
  \caption{Data efficiency of the selector on Terminal-Bench 2.0 (Kimi K2.6, GLM$\rightarrow$Kimi). The selector is trained on a random fraction of tasks and evaluated on all 89. Resolution rate (\%) at $k{=}3$ with the fixed 30-feature set.}
  \label{tab:tts-data-efficiency}
  \footnotesize
  \setlength{\tabcolsep}{8pt}
  \begin{tabular}{l c c c}
    \toprule
    Training fraction & 256K & 32K & 16K \\
    \midrule
    25\%   & 64.0 & 65.2 & 68.5 \\
    50\%   & 66.3 & 64.0 & 68.5 \\
    75\%   & 65.2 & 66.3 & 68.5 \\
    100\%  & 64.0 & 65.2 & 69.7 \\
    \bottomrule
  \end{tabular}
\end{table}